\RequirePackage{fix-cm}
\documentclass[twocolumn]{svjour3}          
\smartqed  
\usepackage{amsmath}
\usepackage{amssymb}
\usepackage{graphicx}
\usepackage{multirow}
\usepackage[linesnumbered,boxed,ruled,commentsnumbered]{algorithm2e}
\begin{document}

\title{Sparse Black-box Video Attack with Reinforcement Learning
}


\author{Xingxing Wei \and Huanqian Yan \and Bo Li
}


\institute{Xingxing Wei \at
Institute of Artificial Intelligence, Hangzhou Innovation Institute, Beihang University, Beijing, China \\
\email{xxwei@buaa.edu.cn}
\and
Huanqian Yan \and Bo Li \at
Beijing Key Laboratory of Digital Media (DML) and State Key Laboratory of Virtual Reality Technology and Systems,\\
School of Computer Science and Engineering, Beihang University, Beijing, China \\
\email{yanhq@buaa.edu.cn, boli@buaa.edu.cn}}

\date{Received: date / Accepted: date}

\maketitle

\begin{abstract}
Adversarial attacks on video recognition models have been explored recently. However, most existing works treat each video frame equally and ignore their temporal interactions. To overcome this drawback, a few methods try to select some key frames and then perform attacks based on them. Unfortunately, their selection strategy is independent of the attacking step, therefore the resulting performance is limited. Instead, we argue the frame selection phase is closely relevant with the attacking phase. The key frames should be adjusted according to the attacking results. For that, we formulate the black-box video attacks into a Reinforcement Learning (RL) framework. Specifically, the environment in RL is set as the recognition model, and the agent in RL plays the role of frame selecting. By continuously querying the recognition models and receiving the attacking feedback, the agent gradually adjusts its frame selection strategy and adversarial perturbations become smaller and smaller. We conduct a series of experiments with two mainstream video recognition models: \emph{C3D} and \emph{LRCN} on the public \emph{UCF-101} and \emph{HMDB-51} datasets. The results demonstrate that the proposed method can significantly reduce the adversarial perturbations with efficient query times.
\end{abstract}
\section{Introduction}
Deep Neural Networks (DNNs) have achieved great success in a wide range of tasks \cite{lecun2015deep,wei2018video,deng2019mixed,silver2017mastering,esteva2017dermatologist,litjens2017a}. Despite this fact,  it is proved that deep neural networks are vulnerable to adversarial examples \cite{goodfellow2014explaining,tramer2018ensemble,dong2019efficient,jia2019comdefend,xie2018mitigating}. Recent works have shown that adding a carefully crafted, small human-imperceptible perturbation to a clean sample can make the deep neural models crash in image classification \cite{akhtar2018threat,goodfellow2014explaining,wei2019transferable}, object detection \cite{zhang2019towards,xie2017adversarial,prakash2018deflecting}, semantic segmentation \cite{xie2017adversarial} and other tasks. Nowadays, more and more DNN models are deployed in various sectors with high-security requirements, which has caused the study of adversarial examples to gain increased attention. The existence of adversarial examples brings huge security risks to the deployment of deep learning systems, such as automatic driving \cite{lu2017no}, robotics \cite{xie2018mitigating}, face recognition \cite{bose2018adversarial,dong2019efficient,gosws2019ijcv} and other aspects.

Due to many real-time video classification systems which are constructed based on the DNN models, it is crucial to investigate the adversarial examples for video models. On the one hand, video attacks can help researchers understand the working mechanism of time-series deep models. On the other hand, adversarial samples facilitate various deep neural network algorithms to assess the robustness by providing more varied video training data \cite{wei2019sparse,dong2019efficient,ddddd2020ijcv}. In this paper, we focus on video attacks, specifically attacking the video classification model under the black-box condition. Compared with the white-box setting, the black-box video attack is more realistic. Because the white-box attack needs to obtain the structures and parameters of the deep learning model, and it is usually difficult in the real applications. In contrast, black-box attacks don't need these information. A widely used way is to access the output of the target model when the input is given. In this way, the gradients can be estimated to generate adversarial examples.

According to \cite{jia2019identifying}, the current video attacking methods can be roughly divided into two classes. The first class is called \emph{dense attack} which pollutes each frame in a video \cite{li2019stealthy,jiang2019black}, and the second class is to select some key frames, and then generates perturbations on these selected frames \cite{wei2019sparse,wei2019heuristic}, called as \emph{sparse attack}. Compared with the dense attack, the sparse attack is more reasonable because there are temporal interactions between adjacent frames in the video. Utilizing this relationship can help both reduce the adversarial perturbations and improve the efficiency of the generation process. For the former advantage, because the selected frames are the most important ones in a video, only adding small perturbations on these frames can fool the recognition model, leading to the reduction of adversarial perturbations on the whole video. For the second advantage, the selected key frames are usually sparse, compared with generating perturbations on the whole frames, the operation of dealing with a few frames is more efficient.

To better select key frames in the sparse attack, a heuristic black-box attack on video recognition models is proposed \cite{wei2019heuristic}. They first propose a heuristic algorithm to evaluate the importance of each frame, and then select key frames by sorting the importance scores, finally, the black-box attacks are performed on the selected frames. However, there are no interactions between the attacking process and the selecting key frames in their method. We argue that key frame selection should not only depend on the video itself, but also on the feedback from the recognition models. The frame selection and video attacking are characterized by mutual guidance and cooperation. The results of this way can produce more accurate key frames and smaller perturbations for adversarial videos.
\begin{figure}[t]
\begin{center}
\includegraphics[width=0.98\linewidth]{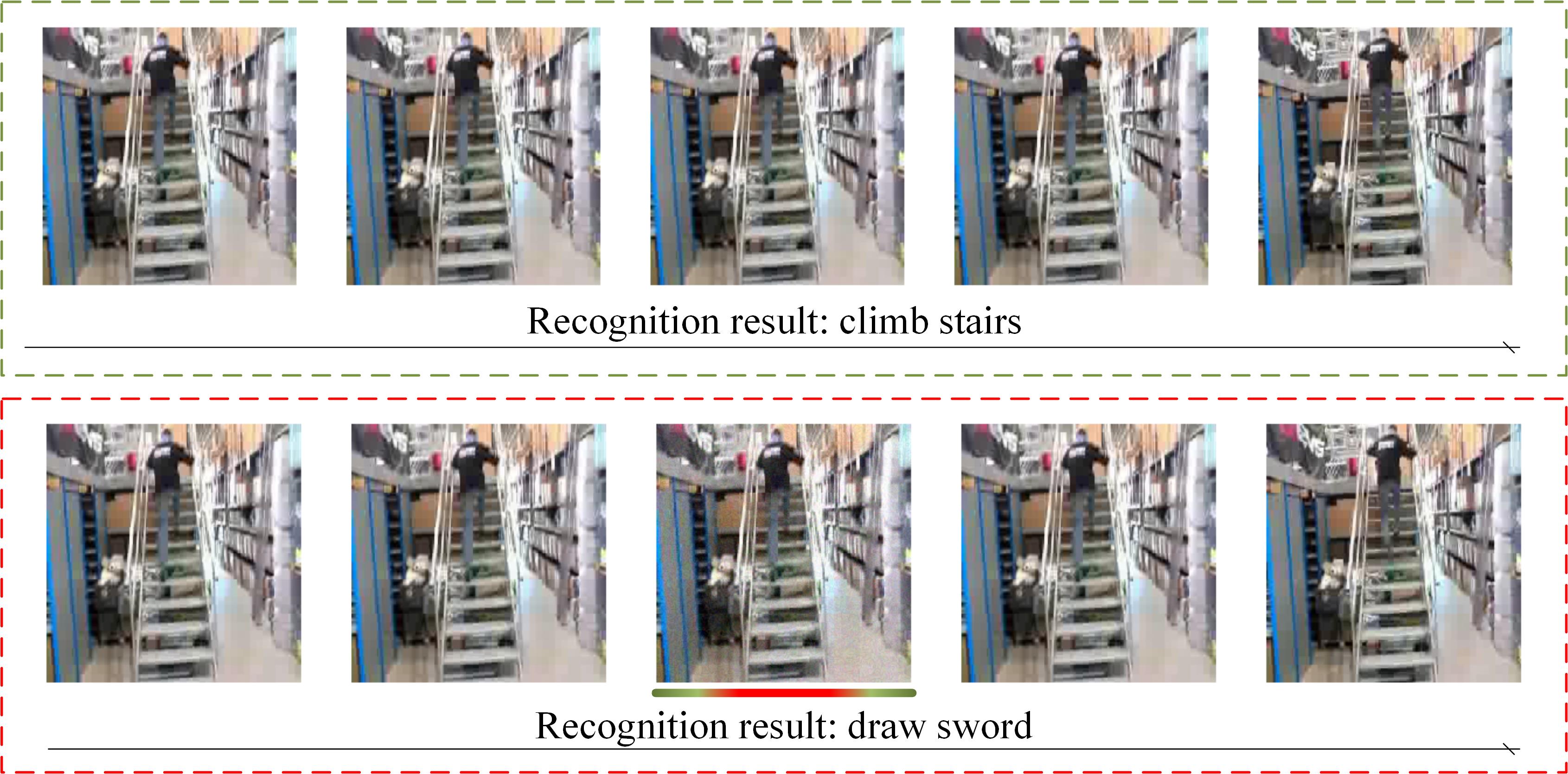}
\end{center}
\caption{An example of sparse black-box video attack with Reinforcement Learning (RL). The clean video (top) can be recognized correctly. The adversarial video (bottom) produced by our proposed method is misclassified. Note that only one frame (green-red line annotation) is adaptively selected by the RL, and very small perturbations are added to the key frame.}
\label{fig:example}
\end{figure}

To this end, we present a Sparse black-box Video Attack (SVA) method with Reinforcement Learning (RL) in this paper. Specifically, the environment in RL is set as the recognition model, and the agent in RL plays the role of frame selecting. By continuously querying the recognition models and receiving the feedback of predicted probabilities (rewards), the agent adjusts its frame selection strategy and performs attacks (actions). Step by step, the optimal key frames are selected and the smallest adversarial perturbations are achieved. 
\begin{figure*}[t]
\begin{center}
\includegraphics[width=0.95\linewidth]{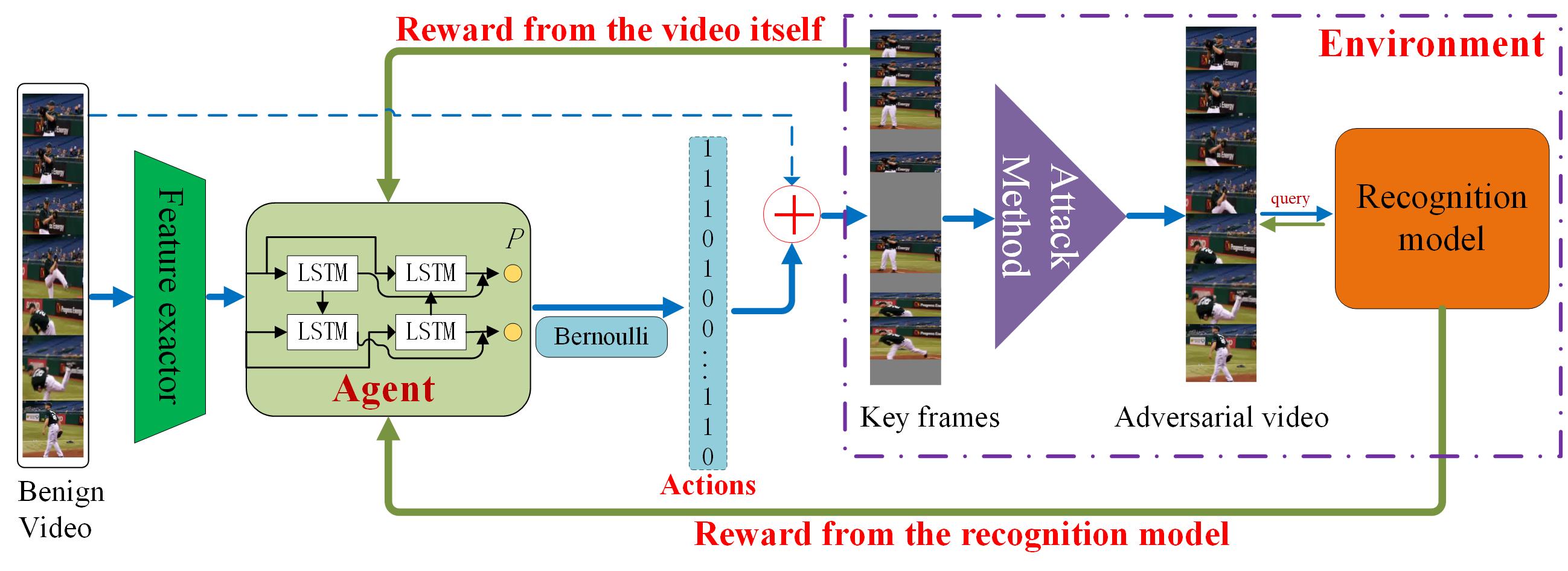}
\end{center}
\caption{Overview of the black-box adversarial video attacking method. We formulate the key frame selection and attacking step into the reinforcement learning framework. Please see the texts for details.}
\label{fig:framework}
\end{figure*}

Technically, for the agent, we use an LSTM-based network \cite{hochreiter1997long} to measure the importance of each frame. 
After obtaining the importance degree $p_t$ of each frame $t$ in the video, Bernoulli sampling is used to determine whether the frames are critical or not. To perform the black-box attacks, the Natural Evolution Strategy (NES) \cite{wierstra2014natural,ilyas2018black} is utilized to estimate the gradient from the recognition models, and then the adversarial videos are generated based on these gradients. For the reward, we design two kinds of functions, the first one comes from the video itself. For example, the frames with big action changes will have the high confidence to be the key frames. The other one comes from the feedback of attacking recognition models.  The insight is that if the frames with tiny perturbations will lead to a big drop of predicted probability, these frames will have the high confidence to be the key frames. Figure \ref{fig:example} shows an adversarial video generated by our proposed method. Figure \ref{fig:framework} overviews the proposed method.  Our major contributions can be summarized as follows:
\begin{itemize}
\item We are the first one to use reinforcement learning to attack video recognition models in the black-box setting. In contrast, previous works perform adversarial attacks on the reinforcement learning model itself, and thus our approach differs fundamentally from those works.
\item A novel algorithm is designed for selecting key frames from a video when attacking video recognition models, which is based on two factors including the visual features of the video itself and the feedback given by the recognition model. Video attacking and key frame selecting are cooperated and guided by each other.
\item Extensive experiments on two widely used video recognition models (LRCN and C3D) and two benchmark video datasets (UCF-101 and HMDB-51) show that the proposed method can significantly reduce the adversarial perturbations while only needing a few query times compared with the state-of-the-art video attacking methods.

\end{itemize}
The rest of this paper is organized as follows. In Section 2, we briefly review the related work. The proposed algorithm is described in Section 3. The experimental results and analysis are presented in Section 4. Finally, we summarize the conclusion in Section 5.

\section{Related Work}
\textbf{Adversarial Attack on Video Models:} Adversarial attacks on images have been extensively studied \cite{guo2017countering,das2018shield,goodfellow2014explaining}. Compared with images, the dimension of videos is very high. It is not easy for general attack algorithms to directly attack such high-dimensional data. A high dimension leads to large search space, and the algorithm needs more query times to find the optimal perturbations to accomplish the successful attack. Therefore, it is more difficult to design an efficient black-box video attack algorithm.

In the past years, many adversarial attacks for videos have been proposed. An $l_{2,1}$-norm regularization based optimization algorithm is the first method that is proposed to compute the sparse adversarial perturbations for video recognition \cite{wei2019sparse}. $l_2$ is used to make the frame have small perturbations and $l_1$ is used to make adversarial frames fewer. The 3D universal perturbation \cite{li2019stealthy} is generated by Generative Adversarial Networks offline and then used with unseen input for the real-time video recognition model. Unlike such white-box attack algorithms which need some knowledge about the video recognition models, Jiang \textit{et al}. utilize tentative perturbations and partition-based rectifications to obtain good adversarial gradient estimates and high attack success rate in the black-box setting \cite{jiang2019black}. But attacking all frames of the video would cause more perturbations and poor robustness of the adversarial video. The adversarial video can easily become ineffective when one of the frames is randomly replaced with a clean video frame. Another black-box method is proposed by Wei \textit{et al}. \cite{wei2019heuristic}, they heuristically search a subset of frames and the adversarial perturbations are only generated on those selected frames, but the attacking processes and key frames selection are separated from each other, the perturbations of adversarial videos are still unsatisfactory. 

Unlike the algorithms mentioned above, our method generates adversarial perturbations on the key frames which are selected by an agent trained using visual features of the video and the feedback of attacking. Our method could generate smaller adversarial perturbations than state-of-the-art black-box video attack methods.

\textbf{Deep Reinforcement Learning:} Deep reinforcement learning is originally designed for learning and mimicking human decision-making processes, which aims to enable the agent to make appropriate behaviors according to the current environment through continuous interaction with the environment \cite{li2017deep,mnih2013playing}. It doesn't require any supervisory information unlike supervised machine learning methods, but rather receives a reward signal to evaluate the performance of the action. Reinforcement learning has received a lot of attention since the AlphaGo \cite{silver2016mastering,silver2017mastering} beats humans. Computer vision tasks have also benefited from deep reinforcement learning in recent years. For example, Zhou \textit{et al}. have applied deep reinforcement learning to train a summarization network for video summary \cite{zhou2018deep}. Dong \textit{et al}. use reinforcement learning for action recognition \cite{dong2019attention}. The process of discarding some irrelevant frames is a kind of hard attention mechanism in their method. Besides, it has been applied in some other fields like tracking, identification \cite{peking2021ijcv}, and person search \cite{li2017deep}.

However, there is no example that deep reinforcement learning is applied in generating adversarial examples. Reinforcement learning algorithms have similar implementation mechanisms with adversarial attack algorithms, especially black-box attack algorithms. It is the first time that we attempt to apply reinforcement learning to the video black-box adversarial attacks. An agent is designed to select key frames while attacking a video using a novel reward function. The key frames selection and adversarial attacks are mutual guidance and cooperation in the whole attacking process.
\section{Methodology}
The adversary takes the video classifier $F(\cdot)$ as a black-box oracle and can only get its output of the top-1 class and its probability. Specifically, given a clean video $x$ and its ground-truth label  $\bar{y}$, $F(\cdot)$ takes  $x$ as an input and outputs the top-1 class label $\textit{F(x)}=y$ and its probability $P(y\vert x)$. If the prediction is correct, then $y=\bar{y}$. The adversarial attack aims to find an adversarial example $x_{adv}$ which can make $F(x_{adv})\neq \bar{y}$ in the un-targeted attack or $F(x_{adv})=y_{adv}$ in the targeted attack with the targeted adversarial class $y_{adv}$, while keeping the adversarial example $x_{adv}$ satisfying the condition: $\parallel x_{adv}-x\parallel_{\rho} \leq \epsilon_{adv}$, where $\epsilon_{adv}$ is the bound of the perturbation $\epsilon$, the $\rho$ in $L_{\rho}$-norm can be set 0,2,$\infty$. 

\subsection{Video Attacking}
The attack algorithm in our method is built based on Fast Gradient Sign Method (FGSM) \cite{goodfellow2014explaining}, which is originally designed for image models. It is defined as:
\begin{equation}
x_{adv} = x +\alpha\cdot sign(\widehat{g}),
\end{equation}
where $\alpha$ is the step size. $sign(\cdot)$ is sign function. $\widehat{g}$ is the gradient, in white-box attacking setting, $\widehat{g}$ can be computed using $\triangledown_{x}l_{adv}(x)$. Here $l_{adv}(x)$ is abbreviated for adversarial loss function, which is described with $l_{adv}(x)=-l(F(x),\bar{y})$ in un-targeted attack and  $l_{adv}(x)= l(F(x),y_{adv})$ in targeted attack. $l(\cdot)$ is a cross-entropy loss. Due to black-box settings, we cannot get the gradient from the recognition model directly, \textit{NES} \cite{wierstra2014natural} algorithm is used as gradient estimator in the proposed method. For \textit{NES} algorithm, it first generates $n/2$ values $\delta_{i}\backsim N(0,I), i\in\lbrace 1,2...n/2\rbrace$, where $n$ is the number of samples. Then, it sets $\delta_{j}=-\delta_{n-j+1}, j\in\lbrace(n/2+1),..n\rbrace$. Finally, the gradient $\widehat{g}$ is estimated as:
\begin{equation}
\widehat{g}\thickapprox \dfrac{1}{\Delta n}\sum_{i=1}^{n} \delta_{i} P(y\vert x+\Delta \cdot \delta_{i}),
\end{equation}
where  $\Delta$ is the search variance.
 
We extend \textit{FGSM} with \textit{NES} from image models to video models as our attacking algorithm. As mentioned above, we use the agent to select the key frames and attack these key frames to achieve the attack of the entire video.  Note that the core contribution in our method is to introduce RL to select key frames, rather than the attacking method module. Therefore, we choose a simple and widely used FGSM+NES method. Other methods like Opt-attack \cite{cheng2019query,cheng2019sign,liu2019signsgd} can also be available.

In addition, for the targeted attack, we first initialize $x_{adv}$ by adding the random noises on the selected frames to make $F(x_{adv})=y_{adv}$. In this time, the perturbation $\epsilon=||x_{adv}-x||_{\rho}$ is large. Next, we will gradually decrease the perturbations' magnitude $\epsilon$ until the given bound $\epsilon_{adv}$ is achieved while keeping the prediction label not change, i.e., the prediction label is still $y_{adv}$. In this way, we obtain the minimal adversarial perturbation for the current selected frames.  
 For the un-targeted attack, we don't need to initialize $x_{adv}$ like the targeted attack, instead, some key frames are randomly selected as victims and the perturbations are added. The key frames will be dynamically adjusted by the agent in the next section, and the optimal perturbation will be solved.

\subsection{Key Frame Selection}
Videos have successive frames in the temporal domain, thus, we consider searching key frames that contribute the most to the success of an adversarial attack. In our approach, key frames selection is considered as a one-step Markov decision process. Figure \ref{fig:framework} provides a sketch map of this process. The agent learns to select the frames by maximizing the total expected reward by interacting with an environment that provides the rewards and updating its actions. 

The input of the agent is a sequence of visual features of the video frames $\lbrace v_{t}\rbrace ^{T}_{t=1}$ with the length \textit{T}. The agent is a bidirectional Long-Short Term Memory network (BiLSTM) \cite{hochreiter1997long} topped with a fully connected (FC) layer. The BiLSTM  produces corresponding hidden states $\lbrace h_{t}\rbrace ^{T}_{t=1}$. We use the ResNet18 \cite{he2016deep} to extract visual features and set the dimension of hidden state in the \textit{LSTM} cell to 128 throughout this paper. Each $h_{t}$ contains both information from the forward hidden state $h^{f}_{t}$ and the backward hidden state $h^{b}_{t}$, which is a good representation of the time domain information of its surrounding frames. The \textit{FC} layer that ends with the sigmoid function $\sigma$ predicts  a probability $p_{t}$ for each frame, and then the key frames $a_{t}$ are sampled via a Bernouli function:
\begin{equation}
p_{t} = \sigma(W\times h_{t}),
\end{equation}
\begin{equation}
a_{t} = Bernoulli(p_{t}),\qquad
\end{equation}
where $a_{t}\in\lbrace 0,1\rbrace$ indicates whether the $t^{th}$ frame is selected or not. $W$ is the weights of the full connection layer. $h_t=\phi(v_t)$, and $\phi(\cdot)$ is the BiLSTM network. 

The reward reflects the quality of different actions. It contains two components in our method: the reward from the inherent attributes of the video itself and the reward from the feedback of the recognition model. The former reward  includes diversity reward $R_{div}$ and representative reward $R_{rep}$ \cite{zhou2018deep}. Let the indices of the selected frames be $K =\lbrace t\vert a_{t}=1,t=1,...,T\rbrace$, the reward $R_{rep}$ and $R_{div}$ can be defined as:
\begin{equation}
R_{rep} = exp(-\dfrac{1}{T} \sum_{t=1}^{T} min_{t^{'}\in K}\Vert v_{t}-v_{t^{'}} \Vert_{2}),
\end{equation}
\begin{equation}
R_{div} = \dfrac{1}{\vert K\vert(\vert K\vert-1)}\sum_{t\in K} \sum_{t^{'}\in K,t^{'}\neq t} d(v_{t},v_{t^{'}}),
\end{equation}
where $d(\cdot,\cdot)$ is the dissimilarity function calculated by
\begin{equation}
d(v_{t},v_{t^{'}})=1-\dfrac{v_{t}^{T}v_{t^{'}}}{\Vert v_{t}\Vert_{2}\Vert v_{t^{'}}\Vert_{2}}.
\end{equation}
The reward from the feedback of the video recognition model is defined as:
\begin{equation}
R_{attack}=\left\{
\begin{array}{lll}
0.999\times exp(\dfrac{-\mathbb{P}}{0.05})  &{30000 > Q > 15000}\\
exp(\dfrac{-\mathbb{P}}{0.05})  &{Q \leqslant 15000}\\
-1  &{Q > 30000},
\end{array} \right.
\end{equation}
where \textit{Q} is the number of queries, $\mathbb{P}$ is the mean perturbation of the adversarial video (MAP), $0.05$ is a normalization factor, $0.999$ is the penalty factor used for reducing the number of queries. The rewards $R_{div}$, $R_{rep}$ and $R_{attack}$ complement each other and work jointly to guide the learning of the agent:
\begin{equation}
R=R_{div}+\gamma_{1}R_{rep}+\gamma_{2}R_{attack}.
\end{equation}
The hyperparameters $\gamma_{1}$ and $\gamma_{2}$ are set according to the parameter tuning.

As shown by the reward function, the reward $R_{rep}$ and the reward $R_{div}$ only rely on the internal properties of the attack video. By the constraints of these two rewards, the agent can accurately identify the video frames that are representative and diverse. For the reward $R_{attack}$, it penalizes the actions versus perturbation and query number meanwhile. For query numbers, we set different rewards in three phases. For queries less than 15000 times, we think the query number is acceptable, and the reward is mainly related with the perturbation $exp(-\mathbb{P}/0.05)$, i.e., if the mean absolute perturbation is small, $exp(-\mathbb{P}/0.05)$ will be large. For queries more than 30000, we think the attack is unavailable, and we align a small reward -1, which is much less than $exp(-\mathbb{P}/0.05)$ for any $\mathbb{P}$. For queries more than 15000 and less than 30000, we add a penalty on $exp(-\mathbb{P}/0.05)$ to make its reward less than that in the first case. In this way, the rewards can encourage the agent to make an accurate decision to achieve the small perturbation and query number at the same time.

Since each frame corresponds to two actions, there are $2^{T}$ possible executions of a video, which is basically not feasible for deep Q learning. Thus, we employ the policy gradient method to make the agent learn a policy function $\pi_{\theta}$ with parameters  $\theta$ by maximizing the expected reward $ J(\theta)=E_{\tau\backsim\pi_{\theta}}[R(\tau)]$. Following the REINFORCE algorithm proposed by Williams \cite{Williams1992Simple}, we approximate the gradient by running the agent for \textit{N} episodes on the same video and then taking the average gradient:
\begin{equation}
\triangledown_{\theta}J(\theta)\approx \dfrac{1}{N} \sum_{n=1}^{N}\sum_{t=1}^{T} R_{n} \triangledown_{\theta} log\pi_{\theta}(a_{t}\vert h_{t}),
\end{equation}
where $R_{n}$ is the reward computed at the $n^{th}$ episode. The number of episodes \textit{N} is set to 5 in our experiments. Although the above method can estimate the gradient well, it may contain high variance which will make the network hard to converge. A common countermeasure is to subtract the reward by a constant baseline \textit{b}, so the gradient becomes
\begin{equation}
\triangledown_{\theta}J(\theta)\approx \dfrac{1}{N} \sum_{n=1}^{N}\sum_{t=1}^{T} (R_{n}-b) \triangledown_{\theta} log\pi_{\theta}(a_{t}\vert h_{t}),
\label{Eq1}
\end{equation}
where \textit{b} is a constant baseline that is used to alleviate the high variance. For computational efficiency, it is set as the moving average of rewards experienced so far. We optimize the policy function's parameter $\theta$ via Adam \cite{kingma2015adam:}.  We update $\theta$ as: $\theta = \theta + ls \cdot\triangledown_{\theta}J(\theta)$, where $ls$ is learning rate and set to $10^{-5}$ in our experiments.
\subsection{Overall Framework}
Here, the whole process of our method in the targeted setting is described in Algorithm \ref{SVA attack algorithm}, which is a continuous-learning algorithm.  The epsilon decay $\triangle_{\epsilon}$ is used to control the reduction size of the perturbation bound. $\triangle_{\epsilon}$ and \textit{FGSM} step size $\alpha$ are dynamically adjusted as described in subsection 3.4. The binary vector \textit{M} has the same size as the frame number of the input video and some values in \textit{M}  will be set to 1 after the agent is trained. The agent in the whole process of attacking is updated with the rewards' values, which is dynamic and interactive. This process does not need any human intervention. 
\begin{algorithm}[t]
  \SetAlgoNoLine 
  \caption{Our SVA targeted attack}  
  \label{SVA attack algorithm}  
  \SetKwInOut{Input}{Input}
  \SetKwInOut{Parameters}{Parameters}
  \SetKwInOut{Output}{Output} 
  \Input{Target class $y_{adv}$ and clean video \textit{x}.}  
  \Output{Adversarial video $x_{adv}$.}
  \Parameters{Maximum $epochs$, episodes $N$ }
  $\lbrace v_{t}\rbrace ^{T}_{t=1} \leftarrow ResNet18(x)$; //extract visual feature \\
  \For{$i=1$ to epochs}{
    $M \leftarrow 0$;\\
    $p_t\leftarrow Agent(\lbrace v_{t}\rbrace ^{T}_{t=1})$ using Eq. (3);\\ 
    \For{$j=1$ to $N$}{
    $a_t \leftarrow Bernoulli(p_{t})$ using Eq.(4);\\
    $M(t) = a_t,t=1,...,T$;\\
    $Q,\mathbb{P},x_{adv} \leftarrow$ Using \textbf{Algorithm} \ref{Targeted video attack};\\
    Compute $R_j \leftarrow$ according to Eq. (9);\\
    }
  	Compute $\triangledown_{\theta}J(\theta) \leftarrow$ according to Eq. (\ref{Eq1});\\
  	Update the \textit{Agent}: $\theta \leftarrow \theta + ls \cdot \triangledown_{\theta}J(\theta)$;\\
  }
 \Return $x_{adv}$ 
\end{algorithm}

\begin{algorithm}[t]
  \SetAlgoNoLine 
  \caption{Video attack algorithm}  
  \label{Targeted video attack}  
  \SetKwInOut{Input}{Input}
  \SetKwInOut{Parameters}{Parameters}
  \SetKwInOut{Output}{Output} 
  \Input{The classifier \textit{F($\cdot$)}, target class $y_{adv}$, clean video \textit{x}, key frame vector $M$.}
  \Output{Mean pixel perturbation $\mathbb{P}$, query numbers $Q$, adversarial video $x_{adv}$.}
  \Parameters{\textit{FGSM} step size $\alpha$, epsilon decay $\triangle_{\epsilon}$, perturbation bound $\epsilon_{adv}$.}
  $x_{adv} \leftarrow$ initializing video with the label $y_{adv}$.\\
  $\epsilon=||x_{adv}-x||_{\rho}$; $Q \leftarrow 0$;\\
  \While{$\epsilon>\epsilon_{adv}$}{
    $\widehat{g} \leftarrow$ according to Eq. (2);\\
    $Q \leftarrow Q+n$; // $n$ is the NES sample numbers.\\
  	$\widehat{g} \leftarrow \widehat{g}\times M$; \\$\widehat{\epsilon}\leftarrow \epsilon-\triangle_{\epsilon}$; //reduce $\epsilon$ value\\ 
  	$x_{adv}^{'} \leftarrow x_{adv}+\alpha\cdot sign(\widehat{g})$ using Eq.(1);\\
  	$\widehat{x}_{adv} \leftarrow Clip(x_{adv}^{'},x-\widehat{\epsilon},x+\widehat{\epsilon})$;\\
  	\eIf{$y_{adv}=F(\widehat{x}_{adv})$}{
  	    $Q \leftarrow Q+1$; $x_{adv} \leftarrow \widehat{x}_{adv}$; $\epsilon \leftarrow \widehat{\epsilon}$;\\  
		}{
		$\widehat{x}_{adv} \leftarrow Clip(x_{adv}^{'},x-\epsilon,x+\epsilon)$.\\
		\If{$y_{adv}=F(\widehat{x}_{adv})$}{
			$Q \leftarrow Q+1$; $x_{adv} \leftarrow \widehat{x}_{adv}$.} 		
  		}
  	Adjust $\triangle_{\epsilon}$ according to the change of $\epsilon$.
  	}
 $\mathbb{P} =\frac{\parallel x_{adv}-x\parallel}{\mid pixel_{x}\mid}
$   // mean pixel perturbation\\
 \Return $\mathbb{P}$, $Q$, $x_{adv}$ 
\end{algorithm}

\subsection{Implementation Details.} 
To follow the query limited black-box settings and make our experiments more convenient, the maximum query number is set to $3\times 10^4$ in the un-targeted mode and $6\times 10^4$ in the targeted mode for all black-box attack algorithms in our experiments. For \textit{NES}, we set the population size as 48, which works well on different datasets in terms of the success rate or the number of queries. For search variance $\Delta$ in \textit{NES}, because the targeted attack needs to keep the target class as the top-1 position, while the un-targeted attack is to remove the current class from the top-1 position, we set it to $10^{-6}$ for the targeted attack setting and $10^{-3}$ for the un-targeted attack setting. The adjustment of step size $\alpha$ adopts the strategy in \cite{madry2017towards}. For the targeted attack, we adjust the step size $\alpha$ and epsilon decay $\vartriangle_{\epsilon}$ dynamically. If the proportion of the adversarial examples cannot be maintained above the threshold 50\%, the step size $\alpha$ is halved. If we can't reduce the perturbation size $\epsilon$ after 20 times in a row, we cut the epsilon decay $\vartriangle_{\epsilon}$ in half.

\begin{figure}[t]
\begin{center}
\includegraphics[width=\linewidth]{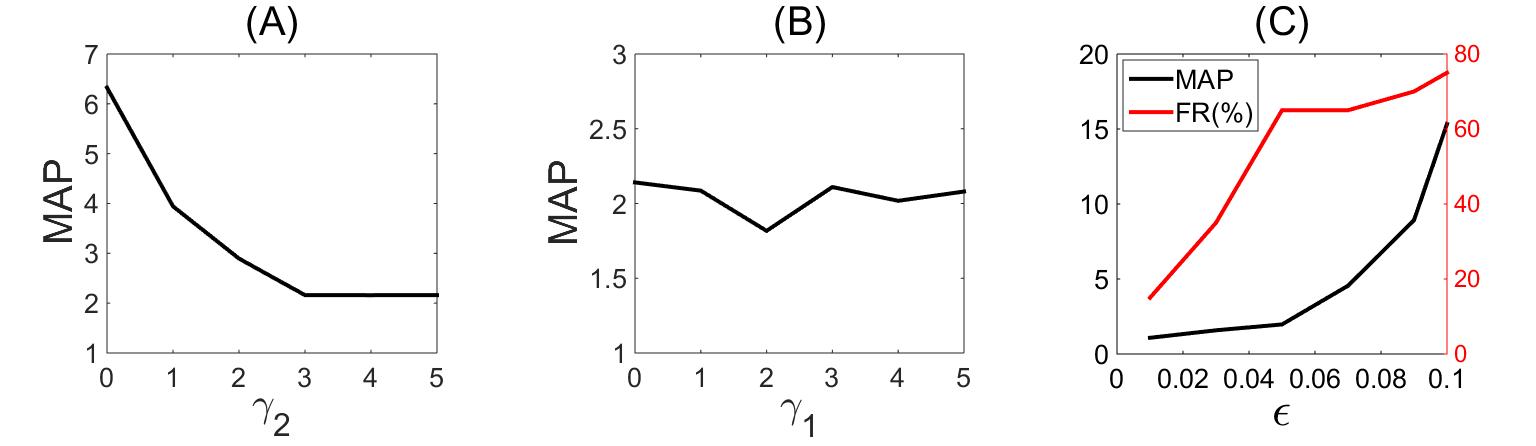}
\end{center}
\caption{Hyperparameters tuning on randomly selected 20 videos and the C3D model with our un-targeted SVA. (A) $\gamma_2$ tuning. (B) $\gamma_1$ tuning. (C) $\epsilon$ tuning.}
\label{fig:gamma}
\end{figure}

There are three hyperparameters in our method, and we obtain their best values via parameter tuning experiments. The $\mathbb{P}$ in Eq.(8) is automatically updated in the algorithm, but we need to set its maximum value $\epsilon_{adv}$.  In Eq.(9), we need tune $\gamma_{1}$ and $\gamma_{2}$.  When $\gamma_{1}$ is tuning, we fix $\gamma_{2}$, and vice versa. The tuning results are given in Figure \ref{fig:gamma}. From the figure, we see $\gamma_{1}=2$ and $\gamma_{2}=3$ are the reasonable choices.  As for $\epsilon_{adv}$, we find that FR (the bigger, the better) and MAP (the lower, the better) reach a balance when $\epsilon=0.05$. Therefore, we set the maximum adversarial perturbations magnitude to $\epsilon_{adv}=0.05$ per frame, and this setting is also applied to other video attack algorithms in the experiment.
\section{Experiments}
In this section, two state-of-the-art video attack models
are used to be compared with our proposed method on two
video recognition models with two public datasets. We focus on
the overall perturbations and the length of the frames selected
in our experiments. Furthermore, a variant of our
method is also designed as a comparison. A comprehensive
evaluation of our method will be presented on those recognition
models and datasets.
\subsection{Datasets and Recognition Models}
The datasets and recognition models used in the experiments are described in detail here.

\textbf{Datasets.} Two widely used datasets, UCF-101 \cite{soomro2012ucf101:} and HMDB-51 \cite{kuehne2011hmdb}, are used in our experiments. UCF-101 is an action recognition dataset containing 13,320 videos with 101 action categories. HMDB-51 is a dataset for human motion recognition, which contains 51 action categories with a total of 7000 videos. Both datasets divide 70\% of the video into training sets and 30\% of the test sets. For convenience and fairness of comparison, 16-frame snippets evenly sampled from each video are used as input to the recognition models during the evaluation. The same approach was fist adopted in \cite{hara2018can} and also used in \cite{wei2019sparse}. In our experiments, we randomly sample 100 videos from the UCF-101 test dataset and 50 videos from the HMDB-51 test dataset for experimental comparison. It is worth noting that all the selected videos can be accurately classified by both recognition models.

\textbf{Recognition Models.} In our experiments, Long-term Recurrent Convolutional Networks (LRCN) \cite{Donahue2014Long} and C3D \cite{hara2018can} are used as recognition models. The LRCN model uses a Recursive Neural Network to encode the Spatio-temporal features generated by CNNs. In our implementation, Inception V3 \cite{szegedy2016rethinking} is used to extract the features of video frames, and LSTM is used for video classification; The C3D model uses 3D convolution to learn Spatio-temporal features from video with Spatio-temporal filter for video classification. These models are all mainstream methods for video classification. In order to make the two video classification models better adapt to the input video, we train them with 16-frame snippets for 30 epochs via Adam in each dataset, and the learning rate is initialized as $10^{-5}$ and decreases to its $1/10$ after 20 epochs. After training, both video classification models meet the requirements of our experiment. Table \ref{accuracy-test} summarizes the test accuracy of the recognition models with 16-frame snippets on the whole UCF101 and HMDB51 datasets.
\begin{table}[t]
\caption{The accuracy of the used video recognition models.} 
\center 
\setlength{\tabcolsep}{7.8mm}{
\begin{tabular}{ccc}
\hline
\multirow{2}{*}{Models} & \multicolumn{2}{c}{Datasets} \\ \cline{2-3} 
 & UCF-101 & HMDB-51 \\ \hline
C3D & 85.88\% & 59.57\% \\ \hline
LRCN & 75.44\% & 34.58\% \\ \hline
\end{tabular}
}
\label{accuracy-test}
\end{table}

\subsection{Evaluation}
We select Fooling rate (FR), Query number (Q), Mean absolute perturbation (MAP), and Sparsity (S) as evaluation metrics to evaluate the performance of our method on various sides. Next, we will introduce and describe these metrics in detail.

\textbf{Fooling rate (FR).} The Fooling rate is defined as the percentage of attacks that successfully generate adversarial examples. In our experiments, it is influenced by two main factors: whether the adversarial video can successfully fool the video classifier and whether it is imperceptible. More concretely, with the upper limit of query numbers ($3\times 10^4$ in the un-targeted mode and $6\times 10^4$ in the targeted mode), if the adversarial video produced by the attack algorithm can make the classifier misclassify itself and the pixel mean perturbations of the whole video is less than the upper limit of perturbations (As described in subsection 3.4, the $\epsilon$ is set to 0.05 in our experiments), the adversarial video is considered successful. Fooling rate (FR) is a common metric for evaluating adversarial attacks, and a higher value means better performance.

\textbf{Query number (Q).} In the query-limited setting, To generate adversarial examples, fewer query times mean that the attack algorithm has a higher attack efficiency. Therefore, we measure the efficiency of the attack algorithm in the current dataset by using the average query number (Q) of all adversarial videos generated with the attack algorithm successfully.

\textbf{Mean absolute perturbation (MAP).} This metric is used to quantify the perceptibility of change in the video. Here, we follow some previous work \cite{Omid2020Pick,2016deepfool,xie2017adversarial} in getting an index (MAP) for an adversarial perturbation given by
\begin{equation}
MAP_{i} =\frac{\parallel x_{i,adv}-x_{i,orig}\parallel}{\mid pixel_{x_i}\mid},  
\end{equation}
where $x_{i,adv}$ is the $i^{th}$ adversarial video, $x_{i,orig}$ is the $i^{th}$ original video, and $\mid pixel_{x_i}\mid$ is the total pixel number of this video. we normalize the $l_1$ norm of the video difference by the total number of pixels. In our experiments, we use the average MAP of all successful adversarial videos on one dataset to evaluate the attack algorithms, and a smaller value means better performance. Additionally, we have resized the value of MAP to 0-255 for the sake of clarity.
\begin{table*}[t]
\center
\caption{The results of SVAL on C3D with UCF-101 under different sparsity (S).}
\setlength{\tabcolsep}{5mm}{
\begin{tabular}{c|c|c|c|c|c|c|c|c}
\hline
\multicolumn{2}{c|}{} & \multicolumn{7}{c}{S(\%)} \\ \hline
 & Metrics & 10 & 20 & 30 & 40 & 50 & 60 & 70 \\ \hline
\multirow{2}{*}{\begin{tabular}[c]{@{}c@{}}Un-targeted\\ Attack\end{tabular}} & MAP & 5.5395 & 5.3805 & 5.3550 & - & \textbf{3.2895} & - & - \\ \cline{2-9} 
 & FR(\%) & 100.0 & 100.00 & 100.00 & 80.0 & \textbf{100.00} & 80.0 & 60.0 \\ \hline
\multirow{2}{*}{\begin{tabular}[c]{@{}c@{}}Targeted\\ Attack\end{tabular}} & MAP & 8.7538 & \textbf{6.6218} & - & - & - & - & - \\ \cline{2-9} 
 & FR(\%) & 100.0 & \textbf{100.0} & 60.0 & 60.0 & 40.0 & 20.0 & 0.0 \\ \hline
\end{tabular}
}
\label{table2}
\end{table*}

\textbf{Sparsity (S).} To show the sparsity effect of the attack algorithms, the sparsity (S), is used to represent the proportion of frames with no perturbations versus all frames in a specific video. It is defined as:
\begin{equation}
S_i=1-\frac{m_i}{T_i},
\end{equation}
where $m_i$ is the length of the key frames of the video $x_i$. A large sparsity value means that only a few frames are added the adversarial perturbations. In our experiments, we use the average $S$ of all successful adversarial videos on one dataset to show the sparse effect of the current attack algorithm.
\subsection{Comparing Algorithms}
We compare our Sparse Video Attack (SVA) method with Opt-attack \cite{cheng2019query} and Heuristic-attack \cite{wei2019heuristic}. For Opt-attack, it is originally proposed to attack image classification models under the black-box setting. The reason we select it as one competitor is that it can achieve smaller distortion compared with some other black-box attack algorithms. We directly extend Opt-attack to attack video models. For Heuristic-attack, it is also a time-domain sparse attack method like ours. We use the same parameter settings in their original papers and official implementation code, respectively.

Besides, one variant of our method, named SVAL, is joined to comparisons. For the SVAL algorithm, we replace the $R_{attack}$ reward in Eq.(9) with the reward $L_{percentage}$ that is used to limit the length of the key frames:
\begin{equation}
L_{percentage} = \Vert\dfrac{1}{T}\sum_{t=1}^{T} p_{t} + S -1\Vert,
\label{Eq2}
\end{equation}
where \textit{S} is the sparsity metric, a bigger \textit{S} value means the fewer frames will be selected. The definition of $p_t$ can be found in Eq.(3). The SVAL algorithm only needs some videos to train the agent and make it intelligent. So in different experiments, we first randomly select 50\% videos to train agent, and then use the agent in our black-box attack method. We still optimize the policy function's parameter $\theta$ via Adam and update $\theta$ as: $\theta = \theta+ls \triangledown_{\theta}(J_{\overline{R_{attack}}}+L_{percentage})$, where $J_{\overline{R_{attack}}}$ means there is no reward $R_{attack}$. The epochs of training is set to 20, and the learning rate is initialized as $10^{-5}$ and decreases to its $1/10$ after 15 epochs.
\subsection{Performance Comparisons with SOTA methods}
Because the parameter \textit{S} in SVAL needs to be set, a grid search method is used to select appropriate parameters with different experiments. We here only show the sparsity tuning for the C3D model with 10 randomly sampled videos from the UCF-101 dataset. The results are recorded in Table \ref{table2}. The symbol ``-" in Table \ref{table2} denotes that the agent cannot achieve the 100\% fooling rate under the current sparsity. In this situation, the MAP cannot be computed across all the adversarial videos, and cannot compare with other sparsity fairly. Therefore, we use ``-" to indicate it.
\begin{table*}[t]
\center
\caption{The video attack results of four attack algorithms in the un-targeted mode. There are four metrics for measuring the effectiveness of the algorithms. For MAP, the smaller the value, the better. For S, the bigger the value, the better. For Q, the smaller the value, the better. For FR, the bigger the value, the better.}
\setlength{\tabcolsep}{4.5mm}{
\begin{tabular}{c|c|l|c|c|c|c}
\hline
\multirow{2}{*}{Dataset} & \multirow{2}{*}{\begin{tabular}[c]{@{}c@{}}Target\\ Model\end{tabular}} & \multirow{2}{*}{Attack Model} & \multicolumn{4}{c}{Metrics \& Un-targeted Attack} \\ \cline{4-7} 
 &  &  & MAP & S(\%) & Q & FR(\%) \\ \hline
\multirow{8}{*}{UCF-101} & \multirow{4}{*}{C3D} & Opt-attack & 4.2540 & 0.00 &15076.23  &74.0  \\ \cline{3-7} 
 &  & Heuristic-attack & 3.2980 & 22.08 & 13609.91 & 79.0\\ \cline{3-7} 
 &  & SVAL(ours) & 3.1765 & 50.00 &\textbf{8367.78}  &83.0  \\ \cline{3-7} 
 &  & SVA(ours) & \textbf{2.4450} & \textbf{63.14} & 9402.28 & \textbf{86.0} \\ \cline{2-7} 
 & \multirow{4}{*}{LRCN} & Opt-attack & 2.8320 & 0.00 &9032.68  &57.0  \\ \cline{3-7} 
 &  & Heuristic-attack & 2.6940 & 17.19 & 9460.38 & 49.0\\ \cline{3-7} 
 &  & SVAL(ours) & 2.4976 & 60.00 & \textbf{4131.57} & \textbf{68.0}  \\ \cline{3-7} 
 &  & SVA(ours) & \textbf{2.396} & \textbf{62.14} & 6132.38 & 63.0 \\ \hline
 \hline
\multirow{8}{*}{HMDB-51} & \multirow{4}{*}{C3D} & Opt-attack & 2.8930 & 0.00 &13274.14  &76.0  \\ \cline{3-7} 
 &  & Heuristic-attack & 2.4960 & 25.68 & 11870.69 & 78.0 \\ \cline{3-7} 
 &  & SVAL(ours) & 2.4482 & \textbf{60.00} &\textbf{10727.93}  &94.0  \\ \cline{3-7} 
 &  & SVA(ours) & \textbf{2.3940} & 51.37 & 24948.67 & \textbf{98.0} \\ \cline{2-7} 
 & \multirow{4}{*}{LRCN} & Opt-attack & 2.7586 & 0.00 & 18207.11 & 62.0  \\ \cline{3-7} 
 &  & Heuristic-attack & 2.6110 & 27.32 & 15663.41 & 66.0 \\ \cline{3-7} 
 &  & SVAL(ours) & \textbf{1.9479} & \textbf{70.00} & \textbf{10891.67} & \textbf{68.0}  \\ \cline{3-7} 
 &  & SVA(ours) & 3.1570 & 62.50 & 18868.09 & 64.0 \\ \hline
\end{tabular}
}
\label{table3}
\end{table*}
\begin{table*}[t]
\center
\caption{The video attack results of four attack algorithms in the targeted mode. There are four metrics for measuring the effectiveness of the algorithms. For MAP, the smaller the value, the better. For S, the bigger the value, the better. For Q, the smaller the value, the better. For FR, the bigger the value, the better.}
\setlength{\tabcolsep}{4.7mm}{
\begin{tabular}{c|c|l|c|c|c|c}
\hline
\multirow{2}{*}{Dataset} & \multirow{2}{*}{\begin{tabular}[c]{@{}c@{}}Target\\ Model\end{tabular}} & \multirow{2}{*}{Attack Model} & \multicolumn{4}{c}{Metrics \& Targeted Attack} \\ \cline{4-7} 
 &  &  & MAP & S(\%) & Q & FR(\%) \\ \hline
\multirow{8}{*}{UCF-101} & \multirow{4}{*}{C3D} & Opt-attack & - & - &$> 60000$  &- \\ \cline{3-7} 
 &  & Heuristic-attack & - & - &$> 60000$  &-  \\ \cline{3-7} 
 &  & SVAL(ours) & 6.7672 & 20.00 &43797.0  & \textbf{38.0} \\ \cline{3-7} 
 &  & SVA(ours) & \textbf{3.6450} & \textbf{57.24} & \textbf{36497.5}  &32.0  \\ \cline{2-7} 
 & \multirow{4}{*}{LRCN} & Opt-attack & - & - &$> 60000$  &-  \\ \cline{3-7} 
 &  & Heuristic-attack & - & - &$> 60000$  &-  \\ \cline{3-7} 
 &  & SVAL(ours) & 5.8834 & 20.00 &\textbf{49065.3}  &39.0  \\ \cline{3-7} 
 &  & SVA(ours) & \textbf{3.270} & \textbf{56.64} &57850.4 &\textbf{41.0}  \\ \hline
 \hline
\multirow{8}{*}{HMDB-51} & \multirow{4}{*}{C3D} & Opt-attack & - & - &$> 60000$  &-  \\ \cline{3-7} 
 &  & Heuristic-attack & - & - &$> 60000$  &- \\ \cline{3-7} 
 &  & SVAL(ours) & 6.9279 & 30.00 &47190.3  &\textbf{40.0}  \\ \cline{3-7} 
 &  & SVA(ours) & \textbf{3.8960} & \textbf{62.15} & \textbf{42900.3}  & 38.0 \\ \cline{2-7} 
 & \multirow{4}{*}{LRCN} & Opt-attack & - & - &$> 60000$  &-  \\ \cline{3-7} 
 &  & Heuristic-attack & - & - &$> 60000$  &-  \\ \cline{3-7} 
 &  & SVAL(ours) & 6.2861 & 20.00 & \textbf{43880.5}  & 32.0 \\ \cline{3-7} 
 &  & SVA(ours) & \textbf{3.5170} & \textbf{66.77} & 47681.9  &\textbf{36.0}  \\ \hline
\end{tabular}
}
\label{table4}
\end{table*}

It is found that the fooling rate decreases with the rising of sparsity. In the un-targeted attack setting, when FR is 100\%, the smallest MAP is 3.2895. Therefore, we set $S = 0.5$ in the following experiment. In the targeted attack setting, $S=0.2$ is a good choice, so the setting is used in the following experiments. In the other sparsity settings, we use the same way to select the corresponding best results. As shown in Table \ref{table2}, it can be found that attacking a part of the video frames is a feasible and effective way, which would significantly reduce the perturbations of the adversarial video. 

The comparison results in the un-targeted setting are listed in Table \ref{table3} in different tasks. In each task, the best performance is emphasized with the bold number. As shown, SVAL and SVA have great advantages over other methods on the whole. On the MAP side, SVA ranks first in the 3/4 comparisons. The biggest gap between other algorithms and SVA occurs in the C3D model with the UCF-101 dataset, the MAP of SVA is only 2.4450, but others all exceed 3. Notice that there is no one case that our methods are not as effective as other methods. On the sparsity side, SVAL and SVA are all ahead of the others, the sparsity generated by them all exceeds 50\%.
\begin{figure*}[t]
\begin{center}
\includegraphics[width=0.80\linewidth]{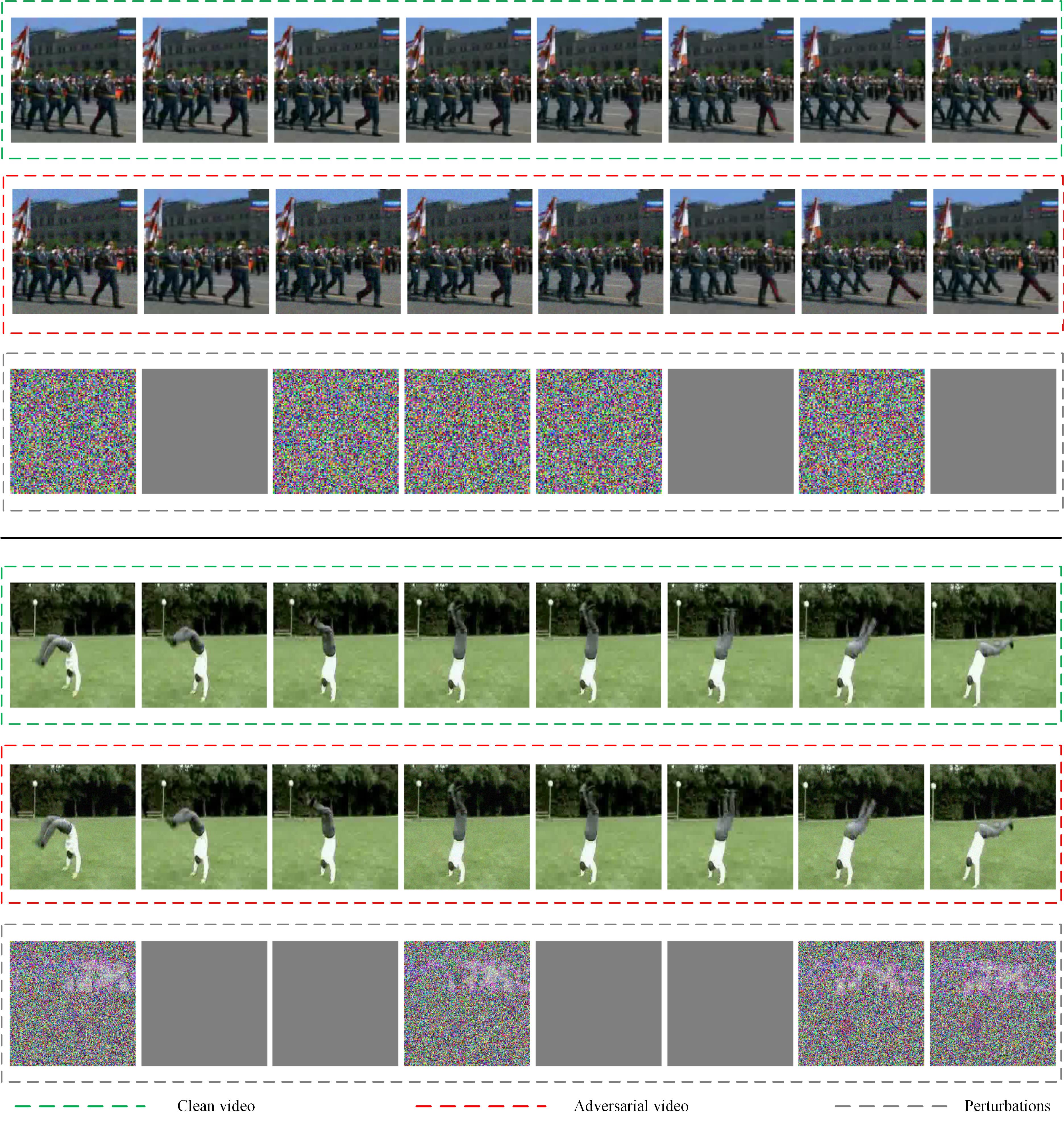}
\end{center}
  \caption{Two adversarial videos produced with SVA un-targeted attack. The clean video, adversarial video, and the corresponding perturbations are shown in the green box, red box, and grey box respectively. The perturbations have been rescaled into the range of 0-255. The adversarial video above the black line is from the UCF-101 dataset with the recognition model C3D, there are 8 frames distributed in video frames 4 to 8, 11, 13, and 14. There are only 5 frames that have the perturbations in the whole 16 frames video. The adversarial video below the black line is from the HMDB-51 dataset with the recognition model LRCN. The example is the front half of the adversarial video which has only 4 frames with perturbations, there are no perturbations on the other frames.}
\label{fig:result}
\end{figure*}
\begin{figure*}[t]
\begin{center}
\includegraphics[width=0.80\linewidth]{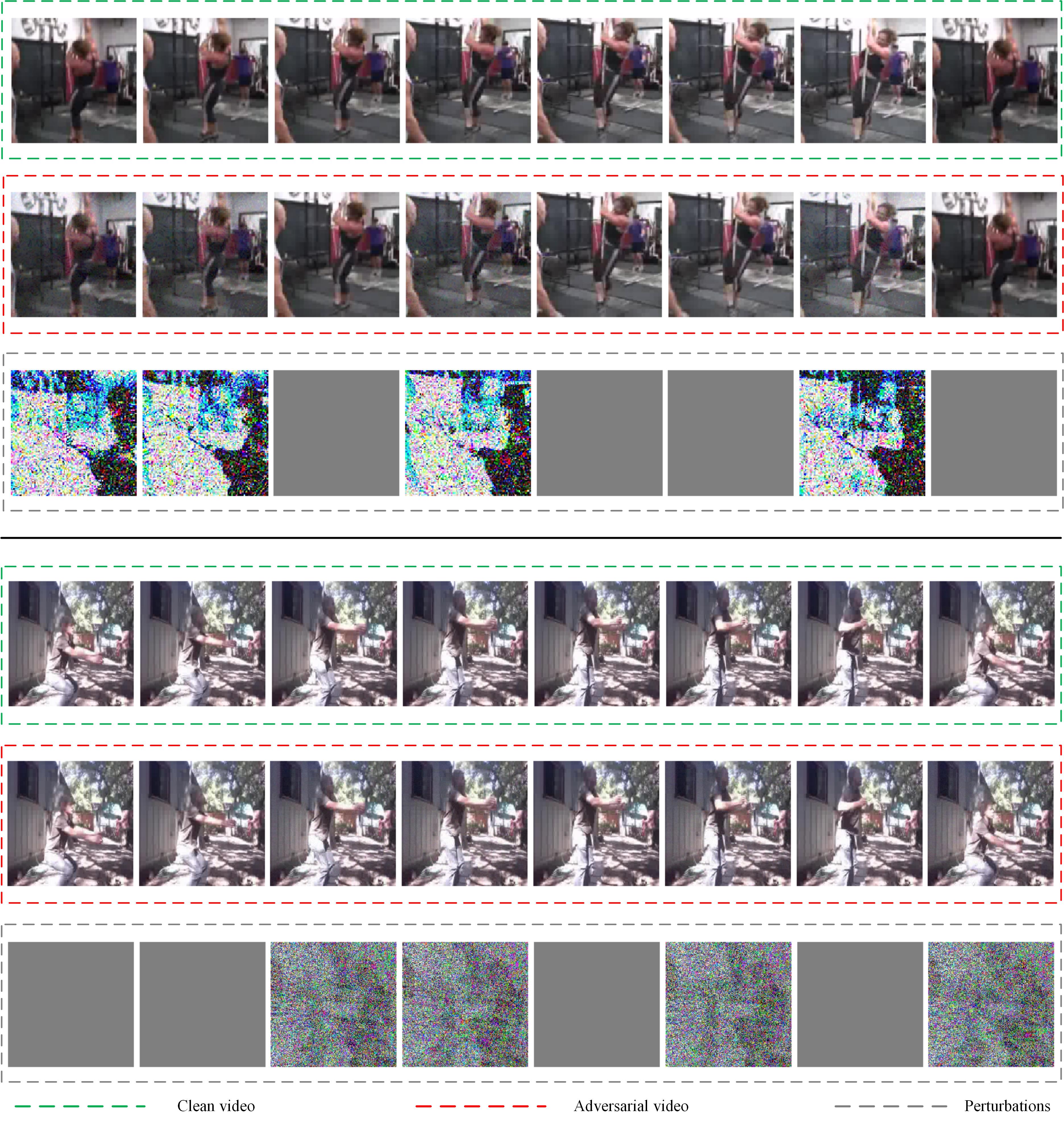}
\end{center}
  \caption{Two examples of the adversarial videos produced with our targeted attack. The clean video, adversarial video, and the corresponding perturbations are shown in the green box, red box, and grey box respectively. The perturbations have been rescaled into the range of 0-255. We display the video frames corresponding to the odd number of their coordinates in two adversarial videos, so only 8 video frames are shown for each video. For these two adversarial videos, the target classes of them are all ’Archery’. The adversarial video above the black line is from UCF-101 with the recognition model C3D, its original class is ’RopeClimbing’. The MAP of this adversarial video is 5.4315, 62.5\% of the total 16 video frames are polluted. The adversarial video below the black line is from UCF-101 with the recognition
model LRCN. Its original class is ’BodyWeightSquats’. The MAP of this adversarial video is 1.479, the perturbations exist in 8 video frames with a total of 16 frames.}
\label{fig:Tresult}
\end{figure*}

Usually, targeted attacks need more query numbers and perturbations than un-targeted attacks. For Opt-attack and Heuristic-attack methods, they don't successfully generate any adversarial video even after 60,000 query times, which exceeds the experimental upper bound of query number pre-defined in section 3.4. Therefore we use ``-" to represent their performance. The comparison results are recorded in Table \ref{table4}. Obviously, the proposed methods are superior to other methods. The FR of our methods is at least 30\% instead of 0\% in the other competitive methods. 

Two examples of the adversarial videos produced with our SVA un-targeted method are shown in Figure \ref{fig:result}. For the first example (above the black line), the ground-truth label is “MilitaryParade”, by adding the generated adversarial perturbations, the model tends to predict a wrong label “BandMarching”. There are only 5 frames that have the perturbations in the whole 16 frames video. For the second example (below the black line), the ground-truth label is “flic flac”, by adding the generated adversarial perturbations, the model tends to predict a wrong label “kick ball”. The adversarial video has only 4 frames with perturbations, there are no perturbations on the other frames. Besides, two examples of the adversarial videos produced with our SVA targeted method are shown in Figure \ref{fig:Tresult}. The clean video, adversarial video, and the corresponding perturbations are shown in the green box, red box, and grey box respectively. For these two adversarial videos, the target classes are all ``Archery". The adversarial video above the black line is from UCF-101 with the recognition model C3D, its original class is ``RopeClimbing". There are 10 video frames that are polluted. The adversarial video below the black line is from UCF-101 with the recognition model LRCN. Its original class is ``BodyWeightSquats". The perturbations only exist in 8 video frames with a total of 16 frames. From these samples, we can conclude that the agent can select a small number of key and representative frames from the whole input video and the perturbations added on these key frames are human-imperceptible.

By observing and analysing all the results, we can draw the following conclusions: (1) Our SVA achieves the best performance in the majority of test tasks. (2) Attacking on key frames is an effective way to reduce perturbations of adversarial video. (3) Mutual guidance and cooperation between key frames selection and attacking is helpful to select the key frames for generating perturbations.
\subsection{Ablation Study}
We conduct a serial of ablation study experiments to analyze the proposed SVA in this subsection. Four experiments are conducted: the effectiveness of different reward functions, the influence of query numbers on experimental results, the effectiveness of keyframe selection algorithms, and the influence of different gradient estimation algorithms on the final attack performance. We discuss the components in the following.
\subsubsection{Different combination of reward functions}
There are three rewards used to guide the agent in our experiments. In this subsection, the proposed method in the un-targeted setting on 20 randomly selected videos with the C3D model is used for the ablation study. The ablation study for each reward function are given in Table \ref{ablation}, where ``No RL" means that the results using FGSM+NES, and no RL module is used, ``SVA$_{R_{attack}}$" means RL module is added, but only $R_{attack}$ is integrated, ``SVA$_{R_{attack+rep}}$" means $R_{attack}$ and $R_{rep}$ are both integrated, and ``SVA" means the full SVA model with $R_{attack}$, $R_{rep}$ and $R_{div}$. It can be shown that the attack reward $R_{attack}$ significantly improves the attacking performance. And the intrinsic rewards $R_{rep}$ and $R_{div}$ have relatively small contributions. The MAP and S in Table \ref{ablation} are computed when FR meets the same accuracy. It can be directly concluded that key frame selection is not only relative to the video itself, but also the feedback from the recognition model. The interactions between the attacking process and the keyframe selection can facilitate successful attacks.
\begin{table}[t]
\center
\caption{The ablation study of the proposed method SVA in an un-targeted setting.}
\setlength{\tabcolsep}{1.6mm}{
\begin{tabular}{c|c|c|c|l}
\hline
\multirow{2}{*}{Metrics} & \multicolumn{4}{c}{Modules} \\ \cline{2-5} 
 & No RL & SVA$_{R_{attack}}$& SVA$_{R_{attack+rep}}$& SVA \\ \hline
MAP & 6.5037 & 2.3723 & 2.0321 & 1.8624 \\ \hline
S(\%) & 0.00 & 62.35 & 68.75 & 74.65 \\ \hline
\end{tabular}
}
\label{ablation}
\end{table}
\subsubsection{Sensitivity to query numbers}
\begin{figure}[t]
\begin{center}
\includegraphics[width=0.95\linewidth]{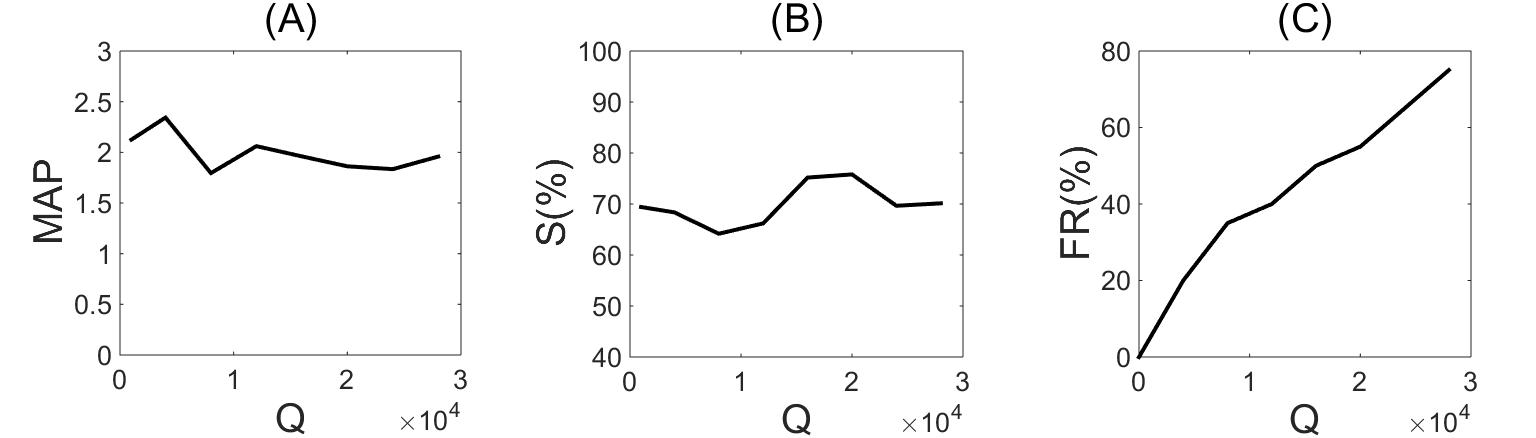}
\end{center}
   \caption{The MAP, S, and FR indices change under the different query number (Q) on randomly selected 20 videos and the C3D model with our un-targeted SVA.}
\label{fig:query}
\end{figure}

We investigate the changes of MAP, S, and FR versus different query numbers (Q). We also run a series of experiments on 20 randomly selected videos and the C3D model using our un-targeted SVA method. The results are shown in Figure \ref{fig:query}. It can be found that FR is very relevant with the query number (Q) while MAP and S are relatively smooth versus the query times. It also indirectly proves that the results in our experiment have certain statistical significance and are relatively robust.
\subsubsection{Effect of keyframe selection}
\begin{figure}[t]
\begin{center}
\includegraphics[width=0.95\linewidth]{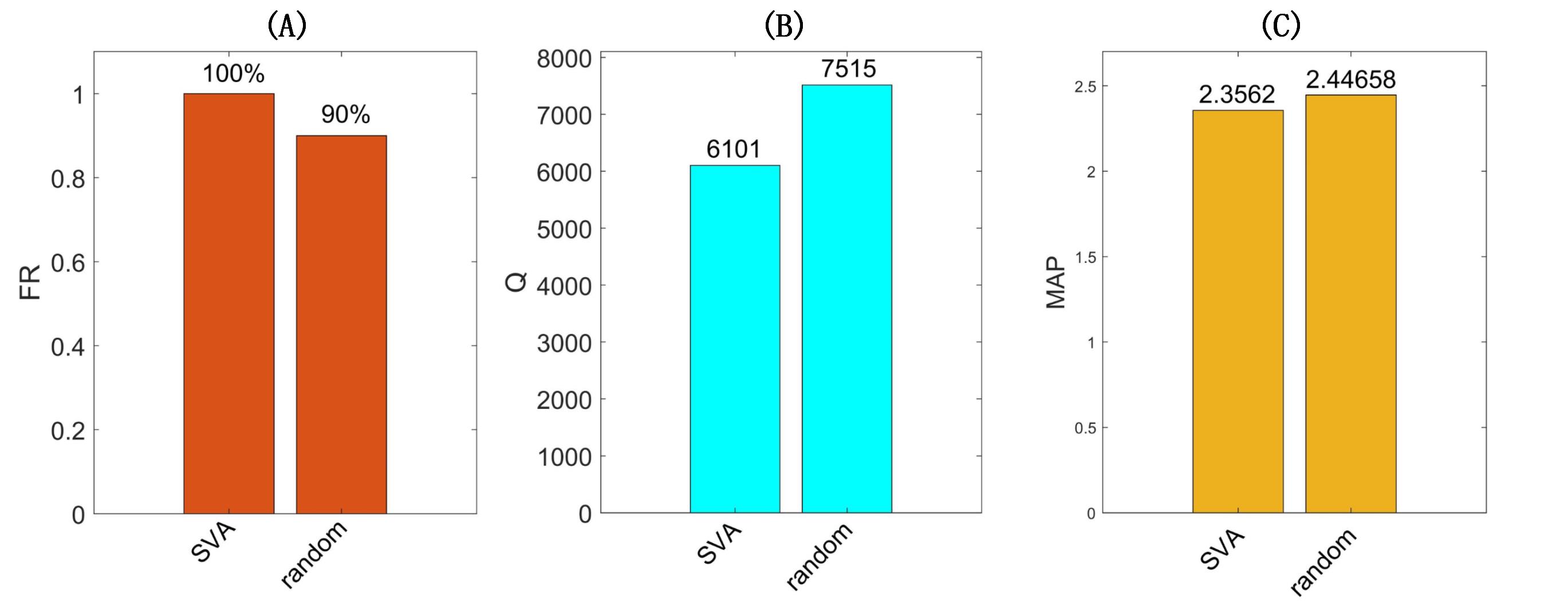}
\end{center}
   \caption{The results of the proposed method SVA and random attack (random) on the C3D recognition model. The only difference between these two methods is the algorithm used to select keyframes. In the experiment, the random algorithm selected the same number of frames as the SVA algorithm, and we recorded the best results in five random attacks on 20 videos from UCF101 datasets. (A) the fooling rate of two algorithms (the bigger the value, the better). (B) the query number of two algorithms (the smaller the value, the better). (C) the MAP values of two algorithms (the smaller the value, the better). Obviously, even the best attack result of the random algorithm is not the same as the SVA algorithm.}
\label{random}
\end{figure}

To illustrate the effectiveness of our keyframe selection algorithm, we make a series of comparisons between our RL-based method and the random selection method. To conduct a fair comparison,  we keep the other components consistent with the SVA and only replace the RL's agent with a random selection step. The number of selected frames is also the same between these two methods. Considering the randomness, each video is attacked five times by the random-based frame selection method, and the best attack result is recorded. The bar chart results of the comparisons are shown in Figure \ref{random}. Obviously, even the best attack result of the random algorithm is not the same as our SVA algorithm.

\begin{figure}[t]
\begin{center}
\includegraphics[width=0.95\linewidth]{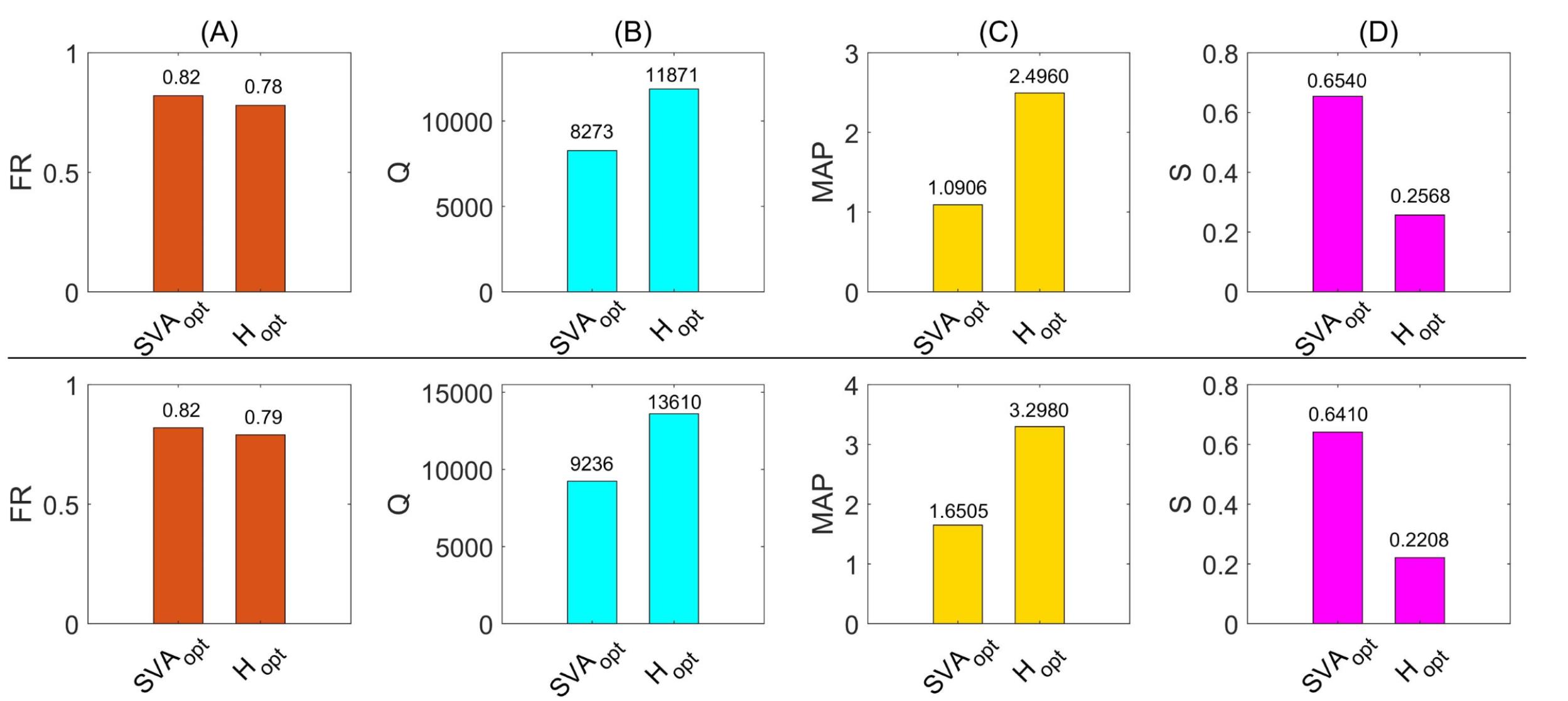}
\end{center}
   \caption{The un-targeted attack results of the Heuristic-attack (H$_{opt}$) and the variant algorithm SVA$_{opt}$ on the C3D recognition model. The proposed keyframe selection method is flexible and can integrate with Heuristic-attack on video attacking (aka SVA$_{opt}$). Here, we use datasets HMDB-51 and UCF-101, the adversarial results above the black line are from HMDB-51, and the adversarial results below the black line are from UCF-101. (A) The fooling rate (FR) of the two methods (the bigger the value, the better). (B) The query number of two methods (the smaller the value, the better). (C) The MAP values of two algorithms (the smaller the value, the better). (D) The sparsity of two algorithms (the bigger the value, the better).}
\label{SVAopt}
\end{figure}

In addition, we also extend our keyframe selection module to the Heuristic-attack \cite{wei2019heuristic} algorithm to replace its frame selection part. The new algorithm (SVA$_{opt}$) and the Heuristic-attack algorithm (H$_{opt}$) have the same parameter settings, they are only different in keyframe selection. Here, we explore the performance of these two algorithms in terms of un-targeted attacks. We use the C3D recognition model, which has a wide application than other video recognition models. The results of these two methods are shown in Figure \ref{SVAopt}. Unsurprisingly, SVA$_{opt}$ has an overwhelming advantage over H$_{opt}$. For both datasets, HMDB-51 and UCF-101, SVA$_{opt}$ leads H$_{opt}$ in all evaluation metrics, which has less perturbation, fewer query numbers, higher sparsity, and higher fooling rate. We can conclude that the proposed keyframe selection algorithm is flexible and efficient.

\subsubsection{Sensitivity to gradient estimation}
\begin{figure}[t]
\begin{center}
\includegraphics[width=\linewidth]{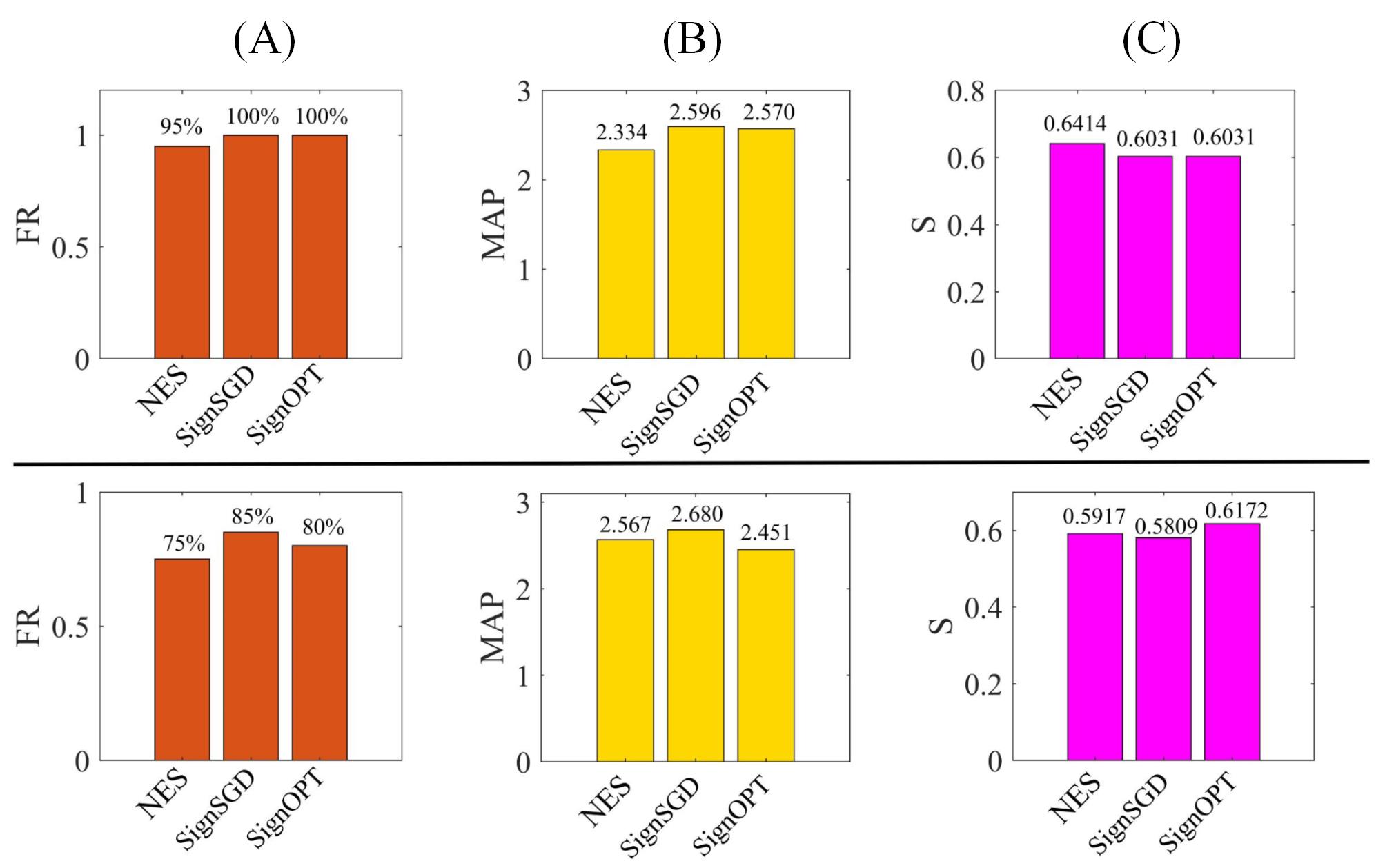}
\end{center}
   \caption{The un-targeted attack results of the proposed SVA on the C3D recognition model with different gradient estimation methods. There are three different gradient estimation algorithms SignSGD, Sign-OPT, and NES. The only different setting between those video attacking methods is the gradient estimation method. Here, we randomly select 20 videos from HMDB-51 and 20 videos from UCF-101, respectively. The adversarial results above the black line are from HMDB-51, and the adversarial results below the black line are from UCF-101. (A) The fooling rate (FR) of the two methods (the bigger the value, the better). (B) The MAP values of two algorithms (the smaller the value, the better). (C) The sparsity of two algorithms (the bigger the value, the better). The max available query times are set to be consistent. It can be concluded from the observation that different gradient estimation methods do not produce significantly different results.}
\label{Grads_ES}
\end{figure}

Here, we make a series of comparisons between different gradient estimation methods, because the gradient estimation method is a key step in our SVA. We test additional gradient methods like SignSGD \cite{liu2019signsgd} and Sign-OPT \cite{cheng2019query} besides the NES method used in our method \cite{wierstra2014natural}. The attacking results of the proposed SVA with three different estimators on the C3D recognition model are shown in Figure \ref{Grads_ES}. Here, we randomly select 20 videos from HMDB-51 and 20 videos from UCF-101, respectively. Each attack algorithm has the same upper query number limitation. From the observation, SVA with different gradient estimators will produce different attack results, but the differences are small. The big difference occurs in the fooling rate. On the UCF101 dataset, SVA with SignSGD and SVA with Sign-OPT obtain 10\% and 5\% better fooling rates than SVA with NES, respectively. For other evaluation metrics, those three methods have no obvious performance advantages. We can conclude that the proposed SVA is not sensitive to the gradient estimation method.
\begin{figure}[t]
\begin{center}
\includegraphics[width=\linewidth]{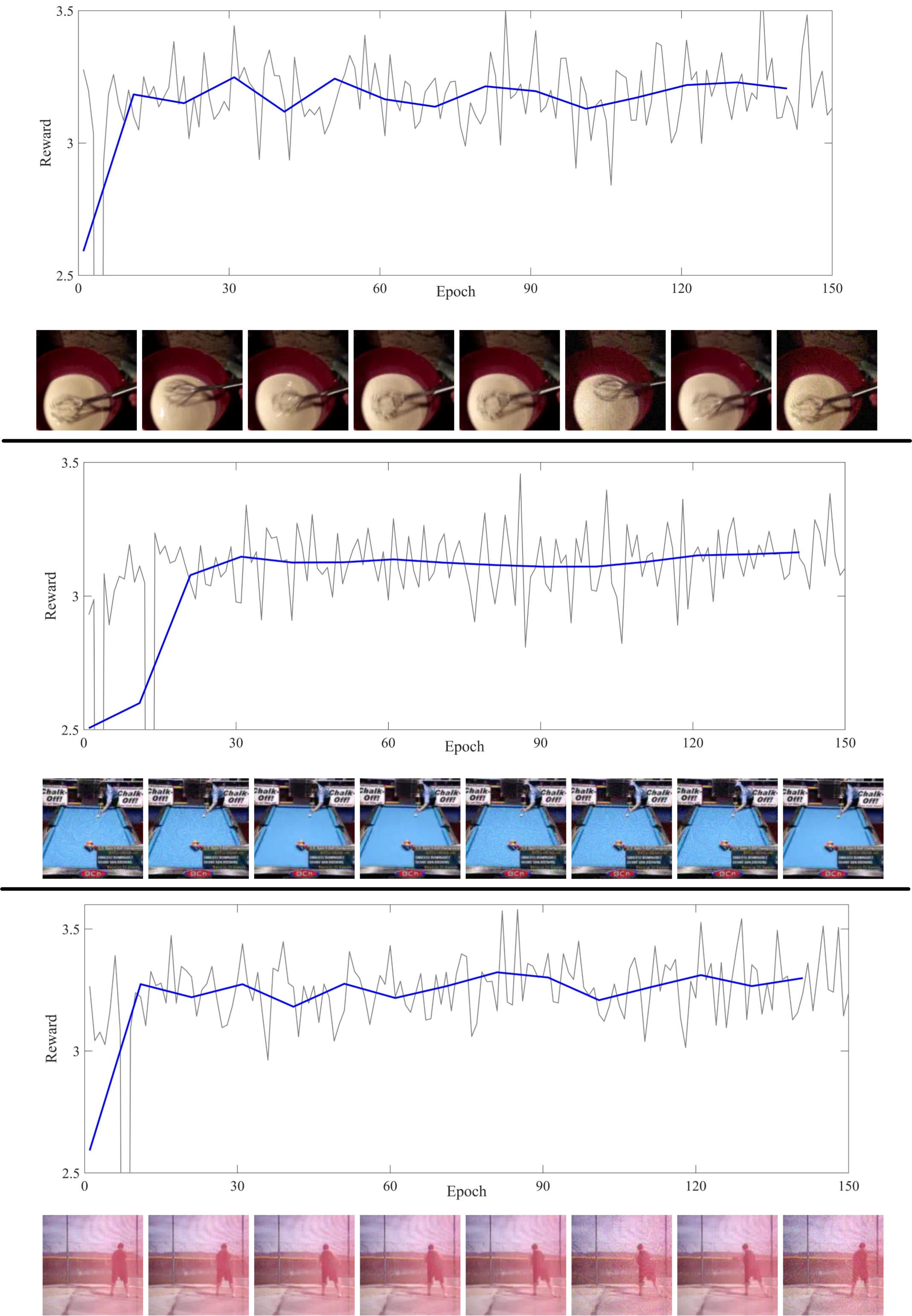}
\end{center}
   \caption{The change of reward value in the different epoch. Here, the C3D recognition model and three randomly selected videos from the UCF101 dataset are used to conduct the experiments. For clear observation, we record each reward value (gray lines) and the corresponding average reward value computed using every 10 epochs (blue lines). The corresponding video frames are also shown below the sub-figures.}
\label{reward}
\end{figure}
\subsection{Convergence of the proposed SVA}
In this subsection, the convergence of the proposed SVA is discussed. We use the change of reward value (i.e., the value computed using Eq.(9)) in different epoch to test whether SVA can obtain a convergence during the learning, which is a widely used way to see the convergence in RL. We conduct experiments under the un-targeted attack setting.  All experiments are based on the C3D model and UCF101 dataset. Due to the limitation of space, we show the change of reward value in the attack process of three separate videos. The reward results are recorded in Figure \ref{reward}. To give a better description, we also show the corresponding video frames (Due to space constraints, only odd-numbered index frames are displayed) below the reward curve. As you can see from the figure, the reward value in each epoch in SVA is choppy (gray lines), but the average reward value per 10 epochs is growing and eventually stabilizing (blue lines). A stable and high reward means that the frame selection strategy is optimal. At the beginning of the attack, attack failure is easy to occur, but this phenomenon will be better in the later stage. The convergence can be achieved after about 20 epochs for these three videos. Actually, this phenomenon is also suitable to other videos. 
\begin{figure}[t]
\begin{center}
\includegraphics[width=\linewidth]{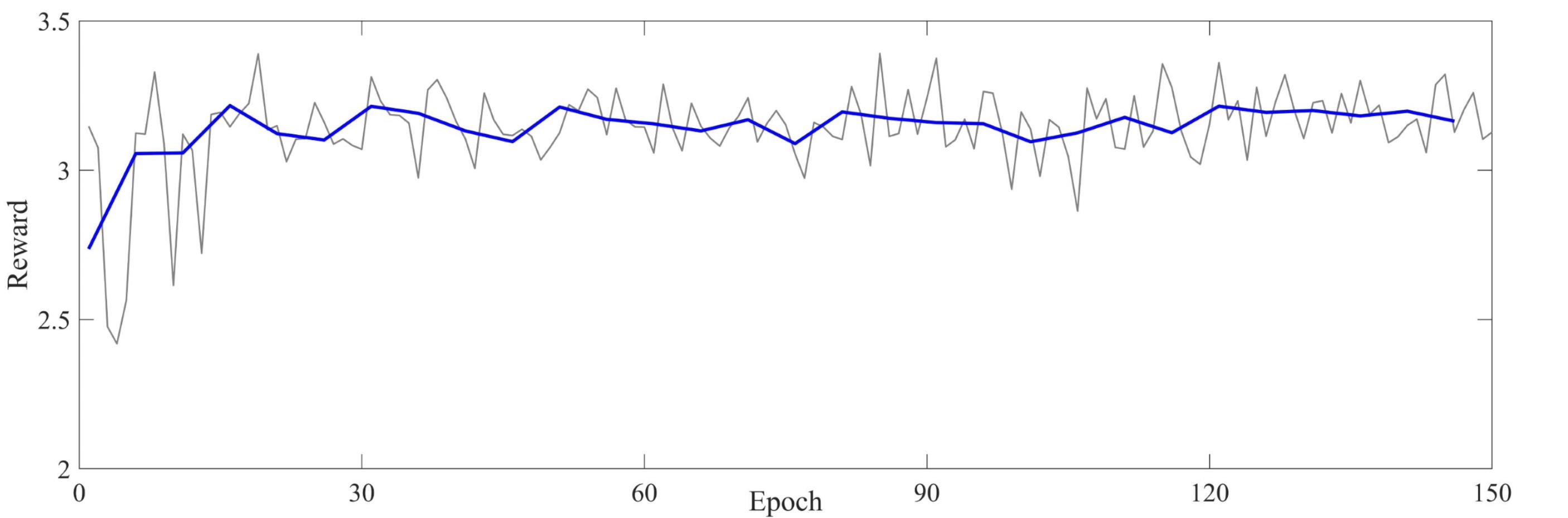}
\end{center}
   \caption{The change of average reward value computed on 20 randomly selected videos. }
\label{reward1}
\end{figure}
We also give the change of average reward computed on 20 randomly selected video attacks. The result is given in Figure \ref{reward1}, where we can see the reward value becomes smooth and stable with the increasing epoch. It shows the good convergence of the proposed SVA.

\section{Conclusion}
In this paper, a sparse black-box adversarial video attack algorithm with reinforcement learning was proposed for video recognition models. Due to a large amount of temporal redundancy information of video data, we explored the sparsity of adversarial perturbations in the video frames through generating adversarial perturbations only on some key video frames. Considering that keyframe selection was not only relevant to the video itself but also the feedback from the recognition model, an agent based on attacking interaction and video intrinsic properties was designed for identifying key frames while attacking. As the perturbations were generated only for the selected frames, the proposed method could reduce the perturbations of adversarial examples significantly. The proposed algorithm was applicable to multiple target models and video datasets. Moreover, the experimental results demonstrated that the proposed algorithm achieved efficient query times to the recognition models. The most pertinent area of future work is to further investigate the black-box attack using fewer queries, such as modifying the update mechanism or designing new rewards. And we hope the query number can be reduced to meet the requirement in the real world.

\section*{Acknowledgments}
This work is supported by National Key R\&D Program of China (Grant No.2020AAA0104002), National Natural Science Foundation of China (No.62076018, 61806109). We also thank anonymous reviewers for their valuable suggestions.


\begin{thebibliography}{}
\bibitem{akhtar2018threat}
Naveed Akhtar and Ajmal Mian.
\newblock Threat of adversarial attacks on deep learning in computer vision: A
  survey.
\newblock {\em IEEE Access}, 6:14410--14430, 2018.

\bibitem{bose2018adversarial}
Avishek~Joey Bose and Parham Aarabi.
\newblock Adversarial attacks on face detectors using neural net based
  constrained optimization.
\newblock pages 1--6, 2018.

\bibitem{cheng2019query}
Minhao Cheng, Thong Le, Pin-Yu Chen, Huan Zhang, Jinfeng Yi, and Cho-Jui Hsieh.
\newblock Query-efficient hard-label black-box attack: An optimization-based
  approach.
\newblock In {\em international conference on learning representations}, 2019.

\bibitem{cheng2019sign}
Minhao Cheng, Simranjit Singh, Pin-Yu Chen, Sijia Liu, and Cho-Jui Hsieh.
\newblock Sign-opt: A query-efficient hard-label adversarial attack.
\newblock In {\em international conference on learning representations}, 2020.

\bibitem{das2018shield}
Nilaksh Das, Madhuri Shanbhogue, Shangtse Chen, Fred Hohman, Siwei Li, Li Chen,
  Michael~E Kounavis, and Duen~Horng Chau.
\newblock Shield: Fast, practical defense and vaccination for deep learning
  using jpeg compression.
\newblock {\em knowledge discovery and data mining}, pages 196--204, 2018.

\bibitem{deng2019mixed}
Lixi Deng, Jingjing Chen, Qianru Sun, Xiangnan He, Sheng Tang, Zhaoyan Ming,
  Yongdong Zhang, and Tat-Seng Chua.
\newblock Mixed-dish recognition with contextual relation networks.
\newblock {\em Proceedings of the 27th ACM International Conference on
  Multimedia}, 2019.

\bibitem{Donahue2014Long}
Jeff Donahue, Lisa~Anne Hendricks, Marcus Rohrbach, Subhashini Venugopalan,
  Sergio Guadarrama, Kate Saenko, and Trevor Darrell.
\newblock Long-term recurrent convolutional networks for visual recognition and
  description.
\newblock {\em IEEE Transactions on Pattern Analysis and Machine Intelligence},
  39(4):677--691, 2017.

\bibitem{dong2019attention}
Wenkai Dong, Zhaoxiang Zhang, and Tieniu Tan.
\newblock Attention-aware sampling via deep reinforcement learning for action
  recognition.
\newblock In {\em national conference on artificial intelligence}, volume~33,
  pages 8247--8254, 2019.

\bibitem{dong2019efficient}
Yinpeng Dong, Hang Su, Baoyuan Wu, Zhifeng Li, Wei Liu, Tong Zhang, and Jun
  Zhu.
\newblock Efficient decision-based black-box adversarial attacks on face
  recognition.
\newblock In {\em computer vision and pattern recognition}, pages 7714--7722,
  2019.

\bibitem{esteva2017dermatologist}
Andre {Esteva}, Brett {Kuprel}, Roberto~A. {Novoa}, Justin~M. {Ko}, Susan~M.
  {Swetter}, Helen~M. {Blau}, and Sebastian {Thrun}.
\newblock Dermatologist-level classification of skin cancer with deep neural
  networks.
\newblock {\em Nature}, 542(7639):115--118, 2017.

\bibitem{goodfellow2014explaining}
Ian~J Goodfellow, Jonathon Shlens, and Christian Szegedy.
\newblock Explaining and harnessing adversarial examples.
\newblock In {\em international conference on learning representations}, 2015.

\bibitem{guo2017countering}
Chuan Guo, Mayank Rana, Moustapha Cisse, and Laurens~Van Der~Maaten.
\newblock Countering adversarial images using input transformations.
\newblock {\em international conference on learning representations}, 2017.

\bibitem{hara2018can}
Kensho Hara, Hirokatsu Kataoka, and Yutaka Satoh.
\newblock Can spatiotemporal 3d cnns retrace the history of 2d cnns and
  imagenet?
\newblock In {\em computer vision and pattern recognition}, pages 6546--6555,
  2018.

\bibitem{he2016deep}
Kaiming He, Xiangyu Zhang, Shaoqing Ren, and Jian Sun.
\newblock Deep residual learning for image recognition.
\newblock In {\em computer vision and pattern recognition}, pages 770--778,
  2016.

\bibitem{hochreiter1997long}
Sepp Hochreiter and Jurgen Schmidhuber.
\newblock Long short-term memory.
\newblock {\em Neural Computation}, 9(8):1735--1780, 1997.

\bibitem{ilyas2018black}
Andrew Ilyas, Logan Engstrom, Anish Athalye, Jessy Lin, Anish Athalye, Logan
  Engstrom, Andrew Ilyas, and Kevin Kwok.
\newblock Black-box adversarial attacks with limited queries and information.
\newblock In {\em international conference on machine learning}, 2018.

\bibitem{jia2019identifying}
Xiaojun Jia, Xingxing Wei, and Xiaochun Cao.
\newblock Identifying and resisting adversarial videos using temporal
  consistency.
\newblock {\em arXiv preprint arXiv:1909.04837}, 2019.

\bibitem{jia2019comdefend}
Xiaojun {Jia}, Xingxing {Wei}, Xiaochun {Cao}, and Hassan {Foroosh}.
\newblock Comdefend: An efficient image compression model to defend adversarial
  examples.
\newblock In {\em 2019 IEEE/CVF Conference on Computer Vision and Pattern
  Recognition (CVPR)}, pages 6084--6092, 2019.

\bibitem{jiang2019black}
Linxi Jiang, Xingjun Ma, Shaoxiang Chen, James Bailey, and Yu-Gang Jiang.
\newblock Black-box adversarial attacks on video recognition models.
\newblock In {\em acm multimedia}, pages 864--872, 2019.

\bibitem{kingma2015adam:}
Diederik~P Kingma and Jimmy Ba.
\newblock Adam: A method for stochastic optimization.
\newblock {\em international conference on learning representations}, 2015.

\bibitem{kuehne2011hmdb}
Hildegard Kuehne, Hueihan Jhuang, Est{\'\i}baliz Garrote, Tomaso Poggio, and
  Thomas Serre.
\newblock Hmdb: a large video database for human motion recognition.
\newblock In {\em 2011 International Conference on Computer Vision}, pages
  2556--2563. IEEE, 2011.

\bibitem{lecun2015deep}
Yann Lecun, Yoshua Bengio, and Geoffrey~E Hinton.
\newblock Deep learning.
\newblock {\em Nature}, 521(7553):436--444, 2015.

\bibitem{li2019stealthy}
Shasha Li, Ajaya Neupane, Sujoy Paul, Chengyu Song, Srikanth~V Krishnamurthy,
  Amit~K Roy-Chowdhury, and Ananthram Swami.
\newblock Stealthy adversarial perturbations against real-time video
  classification systems.
\newblock In {\em network and distributed system security symposium}, 2019.

\bibitem{li2017deep}
Yuxi Li.
\newblock Deep reinforcement learning: An overview.
\newblock {\em arXiv: Learning}, 2017.

\bibitem{litjens2017a}
Geert J.~S. {Litjens}, Thijs {Kooi}, Babak~Ehteshami {Bejnordi}, Arnaud
  Arindra~Adiyoso {Setio}, Francesco {Ciompi}, Mohsen {Ghafoorian}, Jeroen A.
  W. M. Van~Der {Laak}, Bram~Van {Ginneken}, and Clara~I. {Sánchez}.
\newblock A survey on deep learning in medical image analysis.
\newblock {\em Medical Image Analysis}, 42:60--88, 2017.

\bibitem{lu2017no}
Jiajun Lu, Hussein Sibai, Evan Fabry, and David Forsyth.
\newblock No need to worry about adversarial examples in object detection in
  autonomous vehicles.
\newblock {\em arXiv: Computer Vision and Pattern Recognition}, 2017.

\bibitem{madry2017towards}
Aleksander Madry, Aleksandar Makelov, Ludwig Schmidt, Dimitris Tsipras, and
  Adrian Vladu.
\newblock Towards deep learning models resistant to adversarial attacks.
\newblock {\em international conference on learning representations}, 2017.

\bibitem{mnih2013playing}
Volodymyr {Mnih}, Koray {Kavukcuoglu}, David {Silver}, Alex {Graves}, Ioannis
  {Antonoglou}, Daan {Wierstra}, and Martin~A. {Riedmiller}.
\newblock Playing atari with deep reinforcement learning.
\newblock {\em arXiv preprint arXiv:1312.5602}, 2013.

\bibitem{2016deepfool}
Seyedmohsen Moosavidezfooli, Alhussein Fawzi, and Pascal Frossard.
\newblock Deepfool: A simple and accurate method to fool deep neural networks.
\newblock {\em computer vision and pattern recognition}, pages 2574--2582,
  2016.

\bibitem{Omid2020Pick}
Omid~Mohamad Nezami, Akshay Chaturvedi, Mark Dras, and Utpal Garain.
\newblock Pick-object-attack: Type-specific adversarial attack for object
  detection.
\newblock 2020.

\bibitem{prakash2018deflecting}
Aaditya {Prakash}, Nick {Moran}, Solomon {Garber}, Antonella {DiLillo}, and
  James {Storer}.
\newblock Deflecting adversarial attacks with pixel deflection.
\newblock In {\em 2018 IEEE/CVF Conference on Computer Vision and Pattern
  Recognition}, pages 8571--8580, 2018.

\bibitem{silver2016mastering}
David {Silver}, Aja {Huang}, Christopher~J. {Maddison}, Arthur {Guez}, Laurent
  {Sifre}, George van~den {Driessche}, Julian {Schrittwieser}, Ioannis
  {Antonoglou}, Veda {Panneershelvam}, Marc {Lanctot}, Sander {Dieleman},
  Dominik {Grewe}, John {Nham}, Nal {Kalchbrenner}, Ilya {Sutskever}, Timothy
  {Lillicrap}, Madeleine {Leach}, Koray {Kavukcuoglu}, Thore {Graepel}, and
  Demis {Hassabis}.
\newblock Mastering the game of go with deep neural networks and tree search.
\newblock {\em Nature}, 529(7587):484--489, 2016.

\bibitem{silver2017mastering}
David {Silver}, Julian {Schrittwieser}, Karen {Simonyan}, Ioannis {Antonoglou},
  Aja {Huang}, Arthur {Guez}, Thomas {Hubert}, Lucas {Baker}, Matthew {Lai},
  Adrian {Bolton}, Yutian {Chen}, Timothy~P. {Lillicrap}, Fan {Hui}, Laurent
  {Sifre}, George van~den {Driessche}, Thore {Graepel}, and Demis {Hassabis}.
\newblock Mastering the game of go without human knowledge.
\newblock {\em Nature}, 550(7676):354--359, 2017.

\bibitem{soomro2012ucf101:}
Khurram Soomro, Amir~Roshan Zamir, and Mubarak Shah.
\newblock Ucf101: A dataset of 101 human actions classes from videos in the
  wild.
\newblock {\em arXiv: Computer Vision and Pattern Recognition}, 2012.

\bibitem{szegedy2016rethinking}
Christian Szegedy, Vincent Vanhoucke, Sergey Ioffe, Jonathon Shlens, and
  Zbigniew Wojna.
\newblock Rethinking the inception architecture for computer vision.
\newblock {\em computer vision and pattern recognition}, pages 2818--2826,
  2016.

\bibitem{tramer2018ensemble}
Florian {Tramèr}, Alexey {Kurakin}, Nicolas {Papernot}, Ian~J. {Goodfellow},
  Dan {Boneh}, and Patrick~D. {McDaniel}.
\newblock Ensemble adversarial training: Attacks and defenses.
\newblock In {\em International Conference on Learning Representations}, 2018.

\bibitem{wei2019transferable}
Xingxing Wei, Siyuan Liang, Ning Chen, and Xiaochun Cao.
\newblock Transferable adversarial attacks for image and video object
  detection.
\newblock {\em international joint conference on artificial intelligence},
  pages 954--960, 2019.

\bibitem{wei2018video}
Xingxing Wei, Jun Zhu, Sitong Feng, and Hang Su.
\newblock Video-to-video translation with global temporal consistency.
\newblock {\em Proceedings of the 26th ACM international conference on
  Multimedia}, 2018.

\bibitem{wei2019sparse}
Xingxing Wei, Jun Zhu, Sha Yuan, and Hang Su.
\newblock Sparse adversarial perturbations for videos.
\newblock In {\em national conference on artificial intelligence}, volume~33,
  pages 8973--8980, 2019.

\bibitem{wei2019heuristic}
Zhipeng Wei, Jingjing Chen, Xingxing Wei, and Jiang Yugang.
\newblock Heuristic black-box adversarial attacks on video recognition models.
\newblock In {\em national conference on artificial intelligence}, 2020.

\bibitem{Williams1992Simple}
R.~J. Williams.
\newblock Simple statistical gradient-following algorithms for connectionist
  reinforcement learning.
\newblock {\em Machine Learning}, 8(3-4):229--256, 1992.

\bibitem{xie2018mitigating}
Cihang Xie, Jianyu Wang, Zhishuai Zhang, Zhou Ren, and Alan~L Yuille.
\newblock Mitigating adversarial effects through randomization.
\newblock {\em international conference on learning representations}.

\bibitem{xie2017adversarial}
Cihang Xie, Jianyu Wang, Zhishuai Zhang, Yuyin Zhou, Lingxi Xie, and Alan~L
  Yuille.
\newblock Adversarial examples for semantic segmentation and object detection.
\newblock {\em international conference on computer vision}, pages 1378--1387,
  2017.

\bibitem{zhang2019towards}
Haichao Zhang and Jianyu Wang.
\newblock Towards adversarially robust object detection.
\newblock {\em international conference on computer vision}, pages 421--430,
  2019.

\bibitem{zhou2018deep}
Kaiyang Zhou, Yu Qiao, and Tao Xiang.
\newblock Deep reinforcement learning for unsupervised video summarization with
  diversity-representativeness reward.
\newblock In {\em national conference on artificial intelligence}, 2018.

\bibitem{liu2019signsgd}
Sijia Liu, Pin~Yu Chen, Xiangyi Chen, and Mingyi Hong.
\newblock Signsgd via zeroth-order oracle.
\newblock In {\em 7th International Conference on Learning Representations,
  ICLR 2019}, 2019.
  
\bibitem{wierstra2014natural}
Daan Wierstra, Tom Schaul, Tobias Glasmachers, Yi Sun, Jan Peters, and
  J{\"u}rgen Schmidhuber.
\newblock Natural evolution strategies.
\newblock {\em The Journal of Machine Learning Research}, 15(1):949--980, 2014.

\bibitem{gosws2019ijcv}
Gaurav Goswami, Akshay Agarwal, Nalini Ratha, Richa Singh, and Mayank Vatsa. \newblock Detecting and mitigating adversarial perturbations for robust face recognition.
\newblock In {\em International Journal of Computer Vision} 127.6 (2019): 719-742.

\bibitem{ddddd2020ijcv}
Francesco Croce, Jonas Rauber, and Matthias Hein.
\newblock Scaling up the randomized gradient-free adversarial attack reveals overestimation of robustness using established attacks. 
\newblock In {\em International Journal of Computer Vision}, 2020, 128(4): 1028-1046.
\bibitem{peking2021ijcv}
Shangzhi Teng, Shiliang Zhang, Qingming Huang, and Nicu Sebe 
\newblock Viewpoint and scale consistency reinforcement for UAV vehicle re-identification.
\newblock {\em International Journal of Computer Vision} 129.3 (2021): 719-735.

\end{thebibliography}
\end{document}